\documentclass[lettersize,journal]{IEEEtran}
\usepackage{amsmath,amsfonts}
\usepackage{algorithmic}
\usepackage{algorithm}
\usepackage{array}
\usepackage[caption=false,font=normalsize,labelfont=sf,textfont=sf]{subfig}
\usepackage{textcomp}
\usepackage{stfloats}
\usepackage{url}
\usepackage{verbatim}
\usepackage{graphicx}
\usepackage{cite}

\usepackage{pifont}        
\usepackage{mathtools}
\usepackage{amssymb}
\usepackage{booktabs}
\usepackage{tabularx} 
\usepackage{adjustbox}  
\usepackage{array} 
\usepackage{graphicx}
\usepackage{multirow}
\usepackage[table,xcdraw]{xcolor}
\usepackage{makecell}

\usepackage{placeins}      
\usepackage{threeparttable} 

\begin{document}

\title{SC-Diff: Semantically Calibrated Diffusion for Visible-to-Infrared Image Translation}
\author{Junyin Zhang, Siyu Huang, Jianxiong Ye, Haowei Gong, Ruicheng Zhang, Deyu Meng,~\IEEEmembership{Member,~IEEE}, Chenqiang Gao,~\IEEEmembership{Member,~IEEE}%
\thanks{Junyin Zhang, Siyu Huang, Jianxiong Ye, Haowei Gong and Chenqiang Gao are with the School of Intelligent Systems Engineering, Shenzhen Campus of Sun Yat-sen University, Shenzhen, Guangdong 518107, P.R. China. Corresponding author: Chenqiang Gao (e-mail: gaochq6@mail.sysu.edu.cn).\\
\hspace*{1em}Ruicheng Zhang is with the Tsinghua Shenzhen International Graduate School, Tsinghua University, Shenzhen, Guangdong 518055, P.R. China.\\
\hspace*{1em}Deyu Meng is with the School of Mathematics and Statistics, Xi'an Jiaotong University, Xi'an, Shaanxi 710049, P.R. China.\\
}}
\markboth{}
{SC-Diff: Semantically Calibrated Diffusion for Visible-to-Infrared Image Translation}


\maketitle

\begin{abstract}
Visible-to-infrared image translation offers a practical way to expand infrared training data using abundant visible images. Diffusion models have emerged as a promising approach to this task due to their strong generative performance. However, existing diffusion-based approaches typically use semantic priors as external conditions, without explicitly regulating token interactions inside the denoising network. This makes it difficult to preserve object locations, shapes, and semantic layouts required for reliable annotation reuse. To address this issue, we propose SC-Diff, a semantically calibrated latent diffusion framework that uses semantic priors for both conditional guidance and internal self-attention calibration. Specifically, a pretrained SAM3 model with predefined text prompts first extracts category-specific semantic masks from visible images. These masks are merged into a semantic map and fused with the visible image to form the input condition. The same semantic map is then converted into token-level semantic labels to calibrate self-attention inside the denoising network. Based on these labels, we propose Semantic-Guided Self-Attention Calibration (SGSC), which adaptively introduces positive biases for same-category query-key pairs. The query-wise calibration strength is determined by the dispersion of attention across semantic categories and the attention assigned to the query's own category. The original attention strengths further modulate the bias so that same-category keys with stronger responses receive greater calibration. This soft calibration reduces cross-category interference while retaining global contextual interactions, thereby improving semantic consistency in the generated infrared images. Extensive experiments demonstrate that SC-Diff improves perceptual generation quality and provides more effective synthetic training data for downstream infrared object detection.
\end{abstract}


\section{Introduction}
\begin{figure}[!t]
  \centering
  \includegraphics[width=1.0\columnwidth]{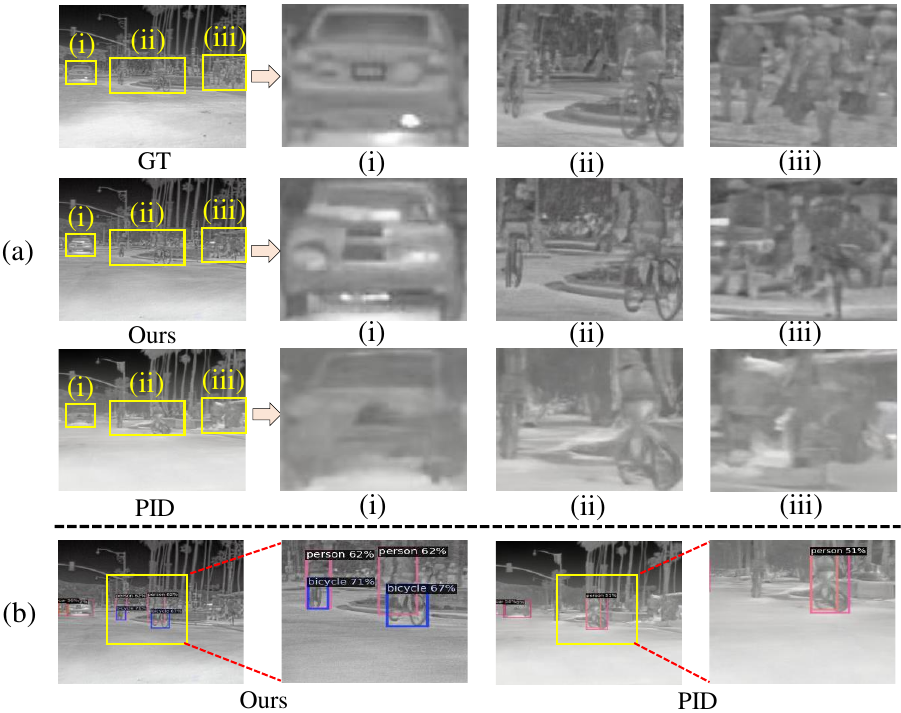}
  \caption{Comparison between PID~\cite{mao2026pid}, a recent representative diffusion-based method, and the proposed SC-Diff on the FLIR~\cite{FLIR_dataset} dataset. (a) Image translation comparison. The first column shows three regions of interest marked by yellow boxes. PID tends to lose fine structures and local details; for example, the bicycle and pedestrian in region (ii) of the third row are largely missing. In contrast, SC-Diff preserves finer structures and stronger local contrast, resulting in more realistic infrared images. (b) Object detection comparison. Using the same detector~\cite{wang2024ov}, structural distortions and missing local details in translated infrared images may lead to missed detections (yellow boxes). By preserving local object structures, SC-Diff leads to fewer missed detections and more complete detection results.}
  \label{fig_1}
\end{figure}
\IEEEPARstart{I}{nfrared}  imaging is widely used in perception tasks because it records thermal radiation and remains reliable under varying illumination. These properties support robust perception in adverse conditions, such as low illumination, smoke, and challenging weather. In particular, infrared object detection has attracted increasing attention in safety-critical applications~\cite{li2023multiscale,2024MMI-Det,EIDet}. 
However, collecting large-scale infrared datasets remains expensive, as it requires dedicated thermal sensors and additional annotation effort. In contrast, visible images are readily available at scale. This makes visible-to-infrared image translation a practical way of expanding infrared training datasets using abundant visible images.

Despite recent progress, generating infrared images that effectively benefit downstream infrared object detection remains challenging. Detection-oriented data augmentation requires not only realistic infrared appearance but also faithful preservation of object positions, shapes, and semantic layouts. Otherwise, structural distortions may invalidate the annotations inherited from visible images and reduce their value for detector training. This makes object-detail preservation particularly important for detection-oriented data augmentation~\cite{2024DPD}. As illustrated in Fig.~\ref{fig_1}, insufficient structural preservation results in missing local details and incomplete object structures.

Visible-to-infrared translation has been explored under different generative paradigms, from adversarial models to recent diffusion-based models. Early adversarial methods can synthesize infrared-like appearances but often suffer from unstable optimization and structural distortions. More recently, diffusion-based methods have shown stronger generation capability and have become a promising solution for visible-to-infrared translation. While diffusion-based methods can generate more realistic infrared appearances, their use of structural or semantic priors is still largely limited to external conditioning. Such priors provide spatial guidance for the generation process, but they do not explicitly regulate token interactions inside the self-attention layers of the denoising U-Net. As a result, unrestricted token aggregation may introduce excessive cross-category interference, leading to weakened object shapes and loss of fine structures, as further examined in Section IV-F.

Motivated by this observation, we propose SC-Diff, a semantically calibrated latent diffusion framework for visible-to-infrared image translation. SC-Diff introduces semantic priors into both condition encoding and self-attention calibration. Specifically, we employ a pretrained SAM3 model~\cite{carion2025sam3} to extract category-specific semantic masks from visible images. These masks are merged into a semantic map, which is combined with the visible image to provide spatial and semantic guidance. Meanwhile, the semantic map is converted into token-level semantic labels and used by the proposed Semantic-Guided Self-Attention Calibration (SGSC) module to regulate token interactions within the denoising U-Net. SGSC adaptively introduces positive biases for same-category token pairs. It uses the original attention weights to emphasize more relevant interactions and enhance semantic consistency. In this way, semantic priors provide external conditional guidance while also calibrating internal feature interactions during denoising.

By improving semantic and structural consistency during translation, SC-Diff makes the annotations inherited from visible images more reliable for reuse. These translated images can serve as effective training samples for infrared object detection, providing a practical way to expand infrared training data and reduce the reliance on large-scale real infrared data collection and manual annotation. The main contributions of this paper are summarized as follows:
\begin{itemize}
    \item We propose SC-Diff, a semantically calibrated latent diffusion framework for visible-to-infrared image translation, aiming to improve semantic and structural consistency for detection-oriented data augmentation.
    \item We introduce SAM3-derived semantic priors and design a Semantic-Guided Self-Attention Calibration (SGSC) module. The module uses token-level semantic labels to enhance same-category token interactions and reduce cross-category feature interference during denoising.
    \item Extensive experiments demonstrate that SC-Diff improves perceptual generation quality and provides more effective augmented data for downstream infrared object detection.
\end{itemize}
The remainder of this paper is organized as follows. Section~\ref{related_work} reviews related work. Section~\ref{method} presents the proposed SC-Diff framework in detail. Section~\ref{experiments} reports the experimental results and analysis. Finally, Section~\ref{conclusion} concludes the paper.
\section{Related Work}
\label{related_work}
In recent years, advances in generative modeling have significantly reshaped image-to-image (I2I) translation. This section first reviews representative methods for general I2I translation, and then discusses recent progress in visible-to-infrared translation.

\subsection{General Image-to-Image Translation}
Generative Adversarial Networks (GANs) have long been a dominant paradigm for image translation. Early supervised methods, such as Pix2Pix~\cite{isola2017image}, use conditional CNNs together with an $L_1$ reconstruction loss to learn pixel-aligned mappings from paired data. To reduce the dependence on paired samples, CycleGAN~\cite{zhu2017unpaired} introduces cycle consistency and enables unsupervised bidirectional translation between two domains. However, when the source and target domains do not form a strictly bijective mapping, unsupervised translation can easily introduce semantic hallucinations or structural changes. StegoGAN~\cite{wu2024stegogan} addresses this issue by using feature-space steganography to disentangle matchable and unmatchable features, thereby improving semantic preservation. PanopticGAN~\cite{PanopticGAN} leverages panoptic priors from foreground instances and background regions to improve semantic and structural consistency during translation. Nevertheless, GAN-based translation may still suffer from unstable optimization and inconsistent structural preservation.

Diffusion models have recently emerged as strong alternatives to GANs for image translation. DDPMs~\cite{ho2020denoising} establish the iterative denoising framework, while LDMs~\cite{rombach2022high} perform diffusion in a compressed latent space to reduce computational cost. Built on the latent diffusion framework, ControlNet~\cite{zhang2023adding} introduces an additional control branch to inject spatial priors, such as depth or edge maps, and enables more precise structural guidance. Other studies further adapt diffusion models to I2I translation. BBDM~\cite{li2023bbdm} models domain translation as a stochastic Brownian bridge connecting two data distributions, while DMT~\cite{xia2024diffusion} improves inference efficiency through single-step domain transfer at an intermediate timestep. AdaNoise~\cite{AdaNoise} uses adaptive noise and dual diffusion with cycle consistency to improve control and preserve semantics. These advances establish diffusion models as flexible frameworks for image translation, while semantic and structural preservation remains particularly important in cross-modal tasks.
\vspace{-10pt}
\subsection{Visible-to-Infrared Image Translation}
Visible-to-infrared image translation aims to convert visible images into infrared representations, providing a low-cost approach to expanding infrared datasets. Early studies predominantly adopted GAN-based frameworks. ThermalGAN~\cite{kniaz2018thermalgan} translates visible person images into thermal representations for cross-modality person re-identification. InfraGAN~\cite{ozkanouglu2022infragan} incorporates reconstruction constraints into a conditional adversarial framework to improve source-target correspondence. EGGAN~\cite{lee2023edge} introduces explicit edge maps into a shared backbone to enhance geometric consistency under limited semantic supervision. Despite these efforts, consistently preserving object structures across diverse scenes remains challenging. Such distortions reduce annotation reliability and limit the utility of generated data for infrared object detection.

More recently, diffusion-based and flow-based methods have shown strong potential for visible-to-infrared image translation due to their generation quality and diversity. Nevertheless, it remains challenging to synthesize realistic infrared appearance while preserving object semantics and local structures from the visible input. To improve generation quality and cross-modal consistency, recent diffusion-based studies introduce different forms of auxiliary guidance. F-ViTA~\cite{paranjape2026f} employs a two-stage pipeline based on Grounding DINO~\cite{liu2024grounding} and SAM~\cite{kirillov2023segment} to establish semantic correspondences across modalities. DiffV2IR~\cite{ran2025diffv2ir} uses progressive learning with a vision-language module and SAM segmentation maps to preserve semantics. PID~\cite{mao2026pid} incorporates physics-informed priors to improve thermal plausibility. TherA~\cite{lee2026thera} employs a thermal-aware vision-language model to generate thermal embeddings for guiding the diffusion process. Beyond diffusion models, ThermalGen~\cite{xiao2025thermalgen} explores style-disentangled flow-based visible-to-infrared translation under varying imaging conditions. Together, these studies highlight the importance of structural guidance and thermal-aware modeling in visible-to-infrared translation.

Despite this progress, recent diffusion-based visible-to-infrared methods mainly introduce semantic priors into the denoising U-Net as external conditions. This strategy provides useful spatial and semantic cues but does not directly regulate token interactions in self-attention. SC-Diff addresses this limitation by applying the same priors to both conditional guidance and self-attention calibration, thereby reducing cross-category interference and better preserving object boundaries and local details.
\section{Methodology}
\label{method}
\begingroup
\setlength{\abovedisplayskip}{4pt plus 1pt minus 1pt}
\setlength{\belowdisplayskip}{4pt plus 1pt minus 1pt}
\setlength{\abovedisplayshortskip}{3pt plus 1pt minus 1pt}
\setlength{\belowdisplayshortskip}{3pt plus 1pt minus 1pt}
This section presents a semantically calibrated diffusion framework for visible-to-infrared image translation in Fig.~\ref{fig_2}.

\subsection{Latent Diffusion Framework}
\label{method_secA}
Visible-to-infrared translation requires the generated infrared image to remain spatially aligned with the visible input, especially in terms of object locations and scene layout. To this end, we build a conditional branch that injects the fused visible-image condition and semantic-map condition into the denoising U-Net at multiple scales. Given a visible image $x^{\text{vis}}$, its infrared counterpart $x^{\text{ir}}$, and the semantic map $x^{\text{seg}}$, a pretrained encoder $\mathcal{E}$ maps them into the latent space:
\begin{equation}
z^{\text{vis}}=\mathcal{E}(x^{\text{vis}}), \qquad
z^{\text{ir}}=\mathcal{E}(x^{\text{ir}}), \qquad
z^{\text{seg}}=\mathcal{E}(x^{\text{seg}}).
\end{equation}
Different from designs that encode the noisy target latent together with the source condition, our condition branch extracts guidance from clean external inputs. The visible latent and semantic latent are first projected to the same feature space and directly fused as:
\begin{equation}
f^{\mathrm{con}}
=
\mathrm{Conv}_{1\times1}(z^{\mathrm{vis}})
+
\mathrm{Conv}_{1\times1}(z^{\mathrm{seg}}).
\end{equation}
The condition encoder further extracts multi-scale guidance features from the fused condition:
\begin{equation}
y_{\mathrm{con}}
=
\{h_{\mathrm{con}}^{(l)}\}_{l=1}^{L}
= c(f^{\mathrm{con}};\theta).
\label{eq_4}
\end{equation}
where $c(\cdot;\theta)$ denotes the condition encoder, $l$ indexes the U-Net layer, and $L$ is the number of guidance layers. The noisy infrared latent is used only as the denoising input, so the condition feature remains clean and is not contaminated by timestep-dependent noise.
Following latent diffusion~\cite{rombach2022high}, the forward corruption process is applied to the target infrared latent $z^{\text{ir}}$:
\begin{equation}
q(z^{\text{ir}}_t \mid z^{\text{ir}}_0)
=
\mathcal{N}\bigl(
z^{\text{ir}}_t;
\sqrt{\bar{\alpha}_t}\, z^{\text{ir}}_0,
(1-\bar{\alpha}_t)\mathbf{I}
\bigr),
\label{eq_1}
\end{equation}
where $z^{\text{ir}}_0=z^{\text{ir}}$, $t$ is the timestep, and $\bar{\alpha}_t=\prod_{s=1}^{t}(1-\beta_s)$. The denoising network $\epsilon_{\phi}$ then predicts the injected noise $\epsilon$ from $z_t^{\text{ir}}$ under the fused clean condition $y_{\mathrm{con}}$. The training objective is:
\begin{equation}
\mathcal{L}_{\text{Diff}}(\phi)
=
\mathbb{E}_{z^{\text{ir}}_0,t,\epsilon}
\bigl[
\|
\epsilon_{\phi}(z_t^{\text{ir}}, y_{\mathrm{con}}, t)-\epsilon
\|^2
\bigr].
\label{eq_2}
\end{equation}
The resulting multi-scale condition features are injected into the corresponding U-Net layers through zero-initialized convolutions. For the $l$-th self-attention layer at diffusion step $t$, the resulting input feature is written as:
\begin{equation}
h_t^{(l)} =
\operatorname{Concat}
\left(
h_{b,t}^{\mathrm{ir},(l)},
h_{s,t}^{\mathrm{ir},(l)} + h_{\mathrm{con}}^{(l)}
\right),
\end{equation}
where $h_{b,t}^{\mathrm{ir},(l)}$ and $h_{s,t}^{\mathrm{ir},(l)}$ denote the backbone feature and skip feature of the denoising U-Net, respectively.

The fused representation $h_t^{(l)}$ is then projected into query, key, and value features:
\begin{equation}
Q_t^{(l)} = W_Q^{(l)} h_t^{(l)}, \quad
K_t^{(l)} = W_K^{(l)} h_t^{(l)}, \quad
V_t^{(l)} = W_V^{(l)} h_t^{(l)} .
\end{equation}
The self-attention map is computed as:
\begin{equation}
A_t^{(l)} =
\operatorname{Softmax}
\left(
\frac{Q_t^{(l)} {K_t^{(l)}}^\top}{\sqrt{d}}
\right),
\end{equation}
where $d$ is the scaling factor. The self-attention output is $\delta^{(l)}_t=A^{(l)}_tV^{(l)}_t$, which updates the decoder representation at diffusion step $t$. In this way, the fused clean condition provides stable multi-scale guidance while avoiding noise leakage from the target latent.

\begin{figure*}[!t]                    
  \centering
  \includegraphics[width=\textwidth]{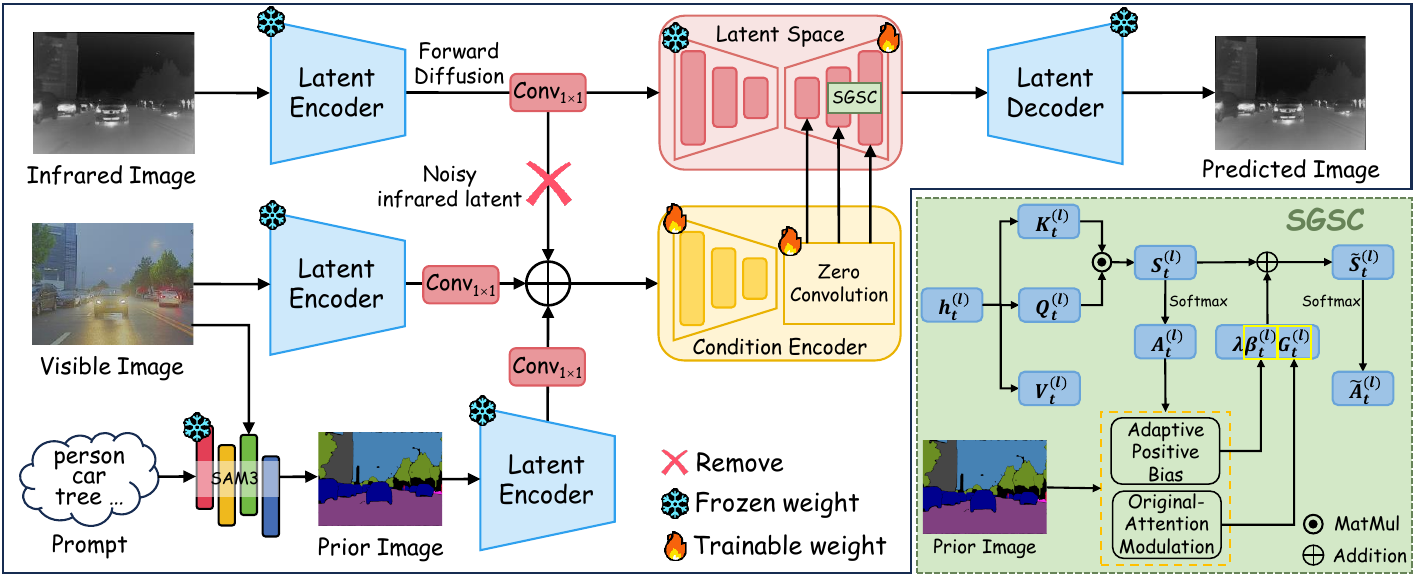} 
 \caption{
 Overall architecture of the proposed SC-Diff. The visible image and semantic map are encoded as clean conditions to provide spatially aligned guidance for the denoising U-Net. The semantic map is obtained from category-specific semantic masks generated by a pretrained SAM3 model with text prompts. During denoising, SGSC uses token-level semantic labels inside the U-Net self-attention layers and adaptively calibrates same-category token interactions. This allows semantic priors to guide both external condition encoding and internal denoising interactions.}
  \vspace{-10pt} 
  \label{fig_2}
\end{figure*}
\vspace{-5pt}
\subsection{Semantic-Guided Self-Attention Calibration}
\label{method_secB}
Although the fused clean condition provides multi-scale guidance to the denoising U-Net, the self-attention layers still aggregate token features without explicit semantic awareness. To regulate these internal interactions, we introduce a two-stage soft calibration strategy using token-level semantic labels inside the U-Net. The first stage computes an adaptive positive bias from the dispersion of attention across semantic categories. The second stage uses the original attention weights to assign larger biases to same-category keys with stronger responses.
\subsubsection{Semantic Mask Generation}
We first use a pretrained SAM3 model, denoted by $\omega$, with predefined text prompts to extract category-specific semantic masks from the visible image. This prompt-based strategy provides category-level regions without requiring manual mask annotations. Given a visible image $x^{\text{vis}}$ and the $k$-th prompt $p_k$, the corresponding category-specific semantic mask is predicted as
\begin{equation}
m_k = \omega(x^{\text{vis}}, p_k),
\end{equation}
where $m_k \in \{0,1\}^{H \times W}$ indicates the region associated with prompt $p_k$. Applying all prompts gives a set of category-specific semantic masks $\{m_k\}_{k=1}^{N}$.

We merge these masks into a semantic label map for attention calibration. When multiple masks overlap, the conflict is resolved using the prediction confidence so that each pixel is assigned to at most one category. Let $\{\bar{m}_k\}_{k=1}^{N}$ denote the resulting non-overlapping masks. The semantic label map $Y$ is defined as:
\begin{equation}
Y(i,j)=
\begin{cases}
k, & \bar{m}_k(i,j)=1,\\
0, & \text{otherwise},
\end{cases}
\end{equation}
where $Y \in \{0,1,\dots,N\}^{H \times W}$ denotes the semantic label map, $Y(i,j)=0$ denotes an unlabeled region that is not assigned to any prompted semantic category, and $Y(i,j)=k$ indicates that pixel $(i,j)$ belongs to the $k$-th category.
\subsubsection{Adaptive Positive Bias}
For the $l$-th self-attention layer at diffusion step $t$, we start from the original attention similarity matrix:
\begin{equation}
S_t^{(l)} = \frac{Q_t^{(l)} K_t^{(l)\top}}{\sqrt{d}} .
\end{equation}
Here, $Q_t^{(l)}$ and $K_t^{(l)}$ are the query and key projections, and $d$ is the feature scaling factor. The original attention map is $A_{t}^{(l)}=\mathrm{Softmax}(S_t^{(l)})$.
 
We then align the semantic label map with the current attention layer. Specifically, $Y$ is resized to the layer resolution and flattened into token-level semantic labels $\boldsymbol{\ell}^{(l)}=\{\ell_i^{(l)}\}_{i\in\Omega_l}$, where $\Omega_l$ denotes the token index set of this layer, $\ell_i^{(l)}\in\{0,1,\dots,N\}$, and $\ell_i^{(l)}=0$ denotes an unlabeled token. For each query token and semantic category $c$, we aggregate the original attention weights assigned to key tokens in that category:
\begin{equation}
P_{t,i,c}^{(l)}
=
\sum_{j\in\Omega_l}
A_{t,ij}^{(l)}
\mathbf{1}
\left[
\ell_j^{(l)}=c
\right],
\label{eq_class_attention_mass}
\end{equation}
where $\mathbf{1}[\cdot]$ is the indicator function. Since label 0 does not represent a semantically coherent category, attention assigned to unlabeled tokens cannot be reliably interpreted as cross-category interference. We therefore exclude this attention mass when determining the calibration strength. Accordingly, $P_{t,i,c}^{(l)}$ is normalized over the prompted semantic categories, and the resulting distribution is denoted by $\widehat{P}_{t,i,c}^{(l)}$. The semantic-category attention entropy score is then computed as:
\begin{equation}
H_{t,i}^{(l)}
=
-\dfrac{1}{\log n^{(l)}}
\displaystyle\sum_{c=1}^{N}
\widehat{P}_{t,i,c}^{(l)}
\log
\left(
\widehat{P}_{t,i,c}^{(l)}+\varepsilon
\right),
\label{eq_attention_entropy}
\end{equation}
where $n^{(l)}$ is the number of prompted semantic categories present at layer $l$, $\varepsilon$ ensures numerical stability, and $H_{t,i}^{(l)}$ is set to zero when $n^{(l)}<2$.

We further consider the proportion of attention assigned to the query token's own semantic category. For a labeled query token, the adaptive positive-bias strength is defined as
\begin{equation}
\beta_{t,i}^{(l)}
=H_{t,i}^{(l)}
\left(1-\widehat{P}_{t,i,\ell_i^{(l)}}^{(l)}\right).
\end{equation}
This formulation assigns stronger calibration when a query distributes its attention across multiple prompted semantic categories while assigning insufficient attention to its own category. Conversely, the entropy term suppresses calibration for highly concentrated attention patterns, thereby avoiding unnecessary modification of confident interactions that may encode useful contextual dependencies. For an unlabeled query token, we set $\beta_{t,i}^{(l)}=0$.
\subsubsection{Original-Attention Modulation}
To restrict the modulation to same-category keys, we construct a same-category indicator matrix:
\begin{equation}
M_{ij}^{(l)}=
\begin{cases}
1, & \ell_i^{(l)} = \ell_j^{(l)},\; \ell_i^{(l)} \neq 0,\; \ell_j^{(l)} \neq 0, \\
0, & \mathrm{otherwise},
\end{cases}
\label{eq_semantic_affinity}
\end{equation}
where $M_{ij}^{(l)}$ indicates whether the $i$-th and $j$-th tokens belong to the same prompted semantic category. Although $\beta_{t,i}^{(l)}$ controls the query-wise calibration magnitude, a uniform logit bias leaves the relative attention distribution among same-category keys unchanged and assigns the same increment to keys with different contextual relevance. Accordingly, we use the original attention weights to modulate the bias:
\begin{equation}
G_{t,ij}^{(l)}
=
\frac{
A_{t,ij}^{(l)}M_{ij}^{(l)}
}{
\max_{r\in\Omega_l} A_{t,ir}^{(l)}M_{ir}^{(l)}+\varepsilon
}.
\label{eq_attention_preserving_gate}
\end{equation}
A scalar factor $\lambda$ controls the overall calibration magnitude and is set to 10. Finally, we obtain the calibrated attention map:
\begin{equation}
\widetilde{A}_{t,ij}^{(l)}
=
\frac{
\exp\left(S_{t,ij}^{(l)}+\lambda\beta_{t,i}^{(l)}G_{t,ij}^{(l)}\right)
}{
\sum_{r\in\Omega_l}
\exp\left(S_{t,ir}^{(l)}+\lambda\beta_{t,i}^{(l)}G_{t,ir}^{(l)}\right)
}.
\label{eq_final_calibrated_attention}
\end{equation}
The calibrated response $\delta_t^{(l)}=\widetilde{A}_{t}^{(l)}V_t^{(l)}$ is then used as the output of the current self-attention layer. We further visualize $A_{t}^{(l)}$ and $\widetilde{A}_{t}^{(l)}$ to compare the self-attention distributions before and after calibration. Details can be found in Section~\ref{ex_vis}.

\subsection{Model Inference}
\label{method_secC}
After training the denoising network $\epsilon_\phi$, inference starts from a Gaussian latent variable $z^{\text{ir}}_T \sim \mathcal{N}(0,\mathbf{I})$. The model then progressively denoises this latent variable under the visible-image condition and semantic-map condition. For a Gaussian reverse transition $p_{\phi}(z^{\text{ir}}_{t-1}\!\mid\! z^{\text{ir}}_{t}, y_{\mathrm{con}})$ with mean $\mu_{\phi}(z^{\text{ir}}_{t},y_{\mathrm{con}},t)$ and covariance $\Sigma_{\phi}(z^{\text{ir}}_{t},y_{\mathrm{con}},t)$, the reverse diffusion step is written as:
\begin{equation}
\begin{aligned}
p_{\phi}(z^{\text{ir}}_{t-1} \mid z^{\text{ir}}_{t}, y_{\mathrm{con}}) 
&= \mathcal{N}\!\big(z^{\text{ir}}_{t-1};\, 
\mu_{\phi}(z^{\text{ir}}_{t}, y_{\mathrm{con}}, t), \\
&\quad \Sigma_{\phi}(z^{\text{ir}}_{t}, y_{\mathrm{con}}, t)\big).
\end{aligned}
\label{eq_9}
\end{equation}
The standard DDPM sampler requires a long Markov chain and is therefore computationally expensive. To improve inference efficiency, we adopt DDIM~\cite{song2020denoising}, which uses a non-Markovian sampling process while preserving the same marginal distributions. The update from step $t$ to $t\!-\!1$ is given by:
\begin{equation}
z^{\text{ir}}_{t-1}
= \sqrt{\bar{\alpha}_{t-1}}\,\hat{z}^{\text{ir}}_0
  + \sqrt{1-\bar{\alpha}_{t-1}-\sigma_t^2}\,\tilde{\epsilon}_{\phi}(z^{\text{ir}}_t,t)
  + \sigma_t \epsilon,
\label{eq_10}
\end{equation}
where $\sigma_t$ controls the stochasticity of the sampling process, and $\epsilon$ denotes Gaussian noise. The clean infrared latent $\hat{z}^{\text{ir}}_0$ is estimated as:
\begin{equation}
\hat{z}^{\text{ir}}_0
= \frac{1}{\sqrt{\bar{\alpha}_t}}\!\left(
z^{\text{ir}}_t - \sqrt{1-\bar{\alpha}_t}\,\tilde{\epsilon}_{\phi}(z^{\text{ir}}_t,t)
\right).
\label{eq_11}
\end{equation}
When $\sigma_t=0$, DDIM sampling becomes deterministic, enabling efficient generation from the initial noise. After the final denoising step, the pretrained decoder $\mathcal{D}$ reconstructs the translated infrared image as $\hat{x}^{\text{ir}} = \mathcal{D}(\hat{z}_0^{\text{ir}})$.

We further apply classifier-free guidance (CFG)~\cite{ho2022classifier} to balance the visible-image condition and the semantic-map condition. During training, we randomly drop the two conditions to enable CFG at inference. Let $y_{\mathrm{con}}^{\varnothing}$, $y_{\mathrm{con}}^{v}$, and $y_{\mathrm{con}}^{v,m}$ denote the condition features obtained by dropping both conditions, keeping only the visible-image condition, and keeping both conditions, respectively. For brevity, we denote their corresponding noise predictions by $\epsilon_{\varnothing}=\epsilon_{\phi}(z_t^{\mathrm{ir}},y_{\mathrm{con}}^{\varnothing},t)$, $\epsilon_v=\epsilon_{\phi}(z_t^{\mathrm{ir}},y_{\mathrm{con}}^{v},t)$, and $\epsilon_{v,m}=\epsilon_{\phi}(z_t^{\mathrm{ir}},y_{\mathrm{con}}^{v,m},t)$. The guided noise estimate is then computed progressively as:
\begin{equation}
\begin{aligned}
\tilde{\epsilon}_{\phi}(z_t^{\mathrm{ir}},t)
=\epsilon_{\varnothing}
+c_v\left(\epsilon_v-\epsilon_{\varnothing}\right)
+c_s\left(\epsilon_{v,m}-\epsilon_v\right),
\end{aligned}
\label{eq_cfg}
\end{equation}
where $c_v$ and $c_s$ denote the guidance scales for the visible-image condition and the semantic-map condition, respectively. The first guidance term introduces spatial structure from the visible image, while the second term further injects semantic region information from the semantic map. When the two guidance scales are equal, the progressive formulation is equivalent to standard CFG applied to the joint visible-image and semantic-map condition. This progressive design allows the denoising process to preserve the input layout and strengthen semantic consistency during infrared generation.
\endgroup
\section{Experiments}
\label{experiments}
\subsection{Experimental Settings}
\begin{table}[!t]
\centering
\caption{Hyperparameters for the proposed SC-Diff. SC-Diff is trained on three NVIDIA A6000 GPUs.}
\label{table_0}
\renewcommand{\arraystretch}{1.2}
\setlength{\tabcolsep}{3.0mm}
\begin{tabular}{llll}
\hline
Hyperparameters  & M$^3$FD & FLIR &KAIST \\ \hline
Input resolution   & 512$\times$512 & 512$\times$512 & 512$\times$512 \\
z shape   & 64$\times$64$\times$4 & 64$\times$64$\times$4 & 64$\times$64$\times$4 \\ 
Diffusion steps   & 1000 & 1000 & 1000 \\ 
Noise schedule   & linear & linear & linear \\ 
Channels   & 320 & 320 & 320 \\ 
Depth   & 1 & 1 & 1 \\ 
Attention resolution   &64,32,16,8 & 64,32,16,8 & 64,32,16,8 \\ 
Channel multiplier   & 1,2,4,4 & 1,2,4,4 & 1,2,4,4 \\ 
Number of heads   & 8 & 8 & 8 \\ 
Batch size per GPU   & 4 & 4 & 4 \\ 
Number of GPUs   & 3 & 3 & 3 \\
Batch accumulation   & 8 & 8 & 8 \\
Epochs     &400  &400  &400   \\ 
Learning rate   & 1e-5 & 1e-5 & 1e-5 \\
Optimizer   & AdamW & AdamW & AdamW \\ 
Pretrained   & SD v1.5 & SD v1.5 & SD v1.5 \\ 
Default sampling steps   & 30 & 30 & 30 \\
DDIM $\eta$   & 0 & 0 & 0 \\ \hline
\end{tabular}
\end{table}
\begin{table*}[htbp]
	\centering
    \renewcommand{\arraystretch}{1.2}
	\caption{Quantitative comparison of generation performance on three public datasets. For PSNR and SSIM, higher is better ($\uparrow$); for LPIPS and FID, lower is better ($\downarrow$). The best results are shown in \textbf{bold}, and the second-best are \underline{underlined}. $\dag$ denotes results quoted from prior publications~\cite{ran2025diffv2ir, mao2026pid} and not reproduced under our evaluation pipeline.}
        \label{table_1}
	\resizebox{\textwidth }{!}{%
	\begin{tabular}{l|cccc|cccc|cccc|c}
	\toprule
	\multirow{2}{*}{\textbf{Method}} & \multicolumn{4}{c|}{\textbf{$\text{M}^3$FD}} & \multicolumn{4}{c|}{\textbf{FLIR}} & \multicolumn{4}{c|}{\textbf{KAIST}} & \multirow{2}{*}{\textbf{Type}} \\ 
	& PSNR $(\uparrow)$ & SSIM $(\uparrow)$ & LPIPS $(\downarrow)$ & FID $(\downarrow)$ & PSNR $(\uparrow)$ & SSIM $(\uparrow)$ & LPIPS $(\downarrow)$ & FID $(\downarrow)$ & PSNR $(\uparrow)$ & SSIM $(\uparrow)$ & LPIPS $(\downarrow)$ & FID $(\downarrow)$\\
	\midrule
	Pix2Pix~\cite{isola2017image}  &18.07 &0.659 &0.353  &160.85 & 17.25 &0.488  &0.396 &150.28 &19.35  &0.714  & 0.408  &169.82 & GAN \\
	CycleGAN~\cite{zhu2017unpaired}  &15.12  &0.593  &0.356  &115.10  &15.54  &0.326  &0.525  &98.70  &17.83  &0.579  &0.425  &74.74  & GAN\\
	ThermalGAN~\cite{kniaz2018thermalgan}&17.89   &0.669  &0.566  &164.29  &17.96 &0.464  &0.512  &147.52  &21.78 &0.750 &0.461 &123.04 & GAN \\
	InfraGAN~\cite{ozkanouglu2022infragan}&10.56  &0.368  &0.567  &340.63  &14.63  &0.389  &0.478  &320.33  &16.45 &0.693 &0.399 &162.21 & GAN\\
	EGGAN-U~\cite{lee2023edge}&12.70   &0.487  &0.342  &125.56   &17.14  &0.371  &0.350  &91.24  &20.06 &0.705 &0.373 &76.27$^\dag$  & GAN\\
	StegoGAN~\cite{wu2024stegogan}&13.97  &0.454  &0.451 &208.56  &13.48  &0.368  &0.543  &103.61 &16.43 &0.553 &0.408 &82.50$^\dag$ & GAN\\
	T2V-DDPM~\cite{nair2023t2v} &10.45  &0.488  &0.461  &162.69  &12.54  &0.453  &0.442  &155.30  &15.40  &0.626 &0.412 &146.58 & Diffusion\\
	F-ViTA~\cite{paranjape2026f}&17.96  &0.681  &0.222  &76.46  &\underline{18.21}  &\textbf{0.499} &0.333  &\underline{82.39}  &19.02 &0.703 &0.323 &68.72  & Diffusion\\
	DiffV2IR~\cite{ran2025diffv2ir} &19.30$^\dag$   &0.662$^\dag$  &\underline{0.200}$^\dag$   &\underline{70.29}$^\dag$   & - & -   & - & - &-  &-  &-  &-  & Diffusion\\
	PID~\cite{mao2026pid} & \underline{20.62}  &\textbf{0.731} &0.214 & 93.85   &18.07  &0.475  &\underline{0.323}  &83.15  &\textbf{24.25} &\textbf{0.810} &\underline{0.255} &\underline{45.13}  & Diffusion\\
	\midrule
		\textbf{SC-Diff (Ours)} &\textbf{20.63} &\textbf{0.731} &\textbf{0.181} &\textbf{62.91} & \textbf{19.15} &\underline{0.496} &\textbf{0.262}  &\textbf{57.66}  &\underline{23.49}  &\underline{0.799}  &\textbf{0.239}  & \textbf{27.98} & Diffusion\\
	\bottomrule
	\end{tabular}
	}
\end{table*}
\begin{figure*}[!t]   
	\centering
	\includegraphics[width=\textwidth]{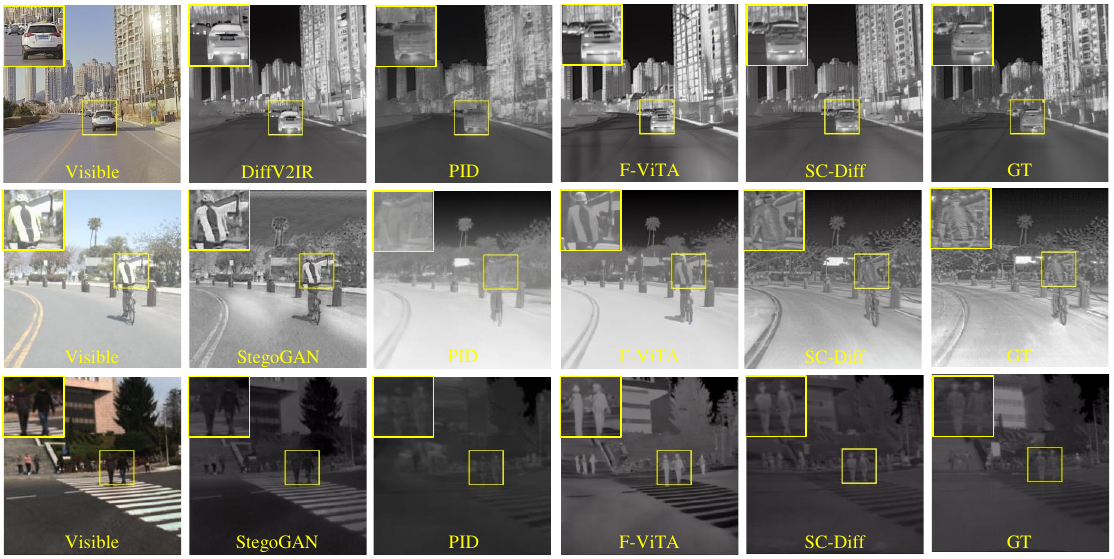}
		\caption{Qualitative comparison under the standard generation setting on three datasets. Each row presents one representative example from a different dataset, and the yellow boxes highlight local regions for zoomed-in comparison.}
	\label{fig_3}
    \vspace{-10pt}
\end{figure*}
\begin{figure*}[t]
	\centering
	\includegraphics[width=\textwidth]{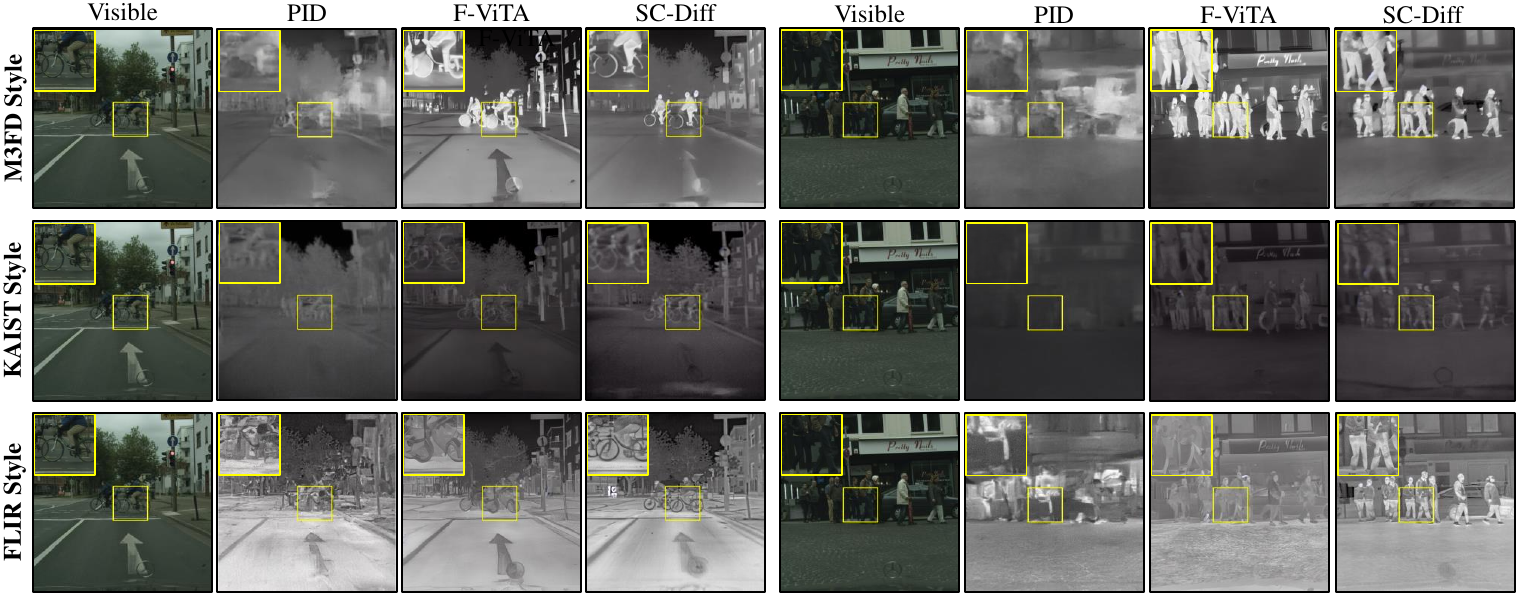}
		\caption{Out-of-distribution generation results on Cityscapes images. Visible images are translated into the infrared styles of M$^3$FD, KAIST, and FLIR. Each group shows one visible scene, with each row corresponding to a specific target infrared style. Yellow boxes highlight representative local regions.}
	\label{fig_4}
\end{figure*}
\noindent \textbf{Datasets.} 
We conduct visible-to-infrared translation experiments on three widely used paired benchmarks, including M$^{3}$FD~\cite{liu2022target}, FLIR~\cite{FLIR_dataset}, and KAIST~\cite{HwangKAIST}. These datasets cover diverse acquisition conditions and scene distributions, allowing us to evaluate in-domain translation performance across different benchmarks. We additionally use Cityscapes~\cite{cordts2016cityscapes} as an external visible-image source to evaluate out-of-distribution generation and the effectiveness of translated images for downstream infrared object detection.

M$^{3}$FD contains 4,200 aligned pairs of visible and infrared images collected from diverse daytime and nighttime scenes. We follow the data split released by DiffV2IR~\cite{ran2025diffv2ir}, using 3,550 pairs for training and 650 pairs for testing. For FLIR and KAIST, we adopt the standard train/test splits commonly used in infrared object detection. FLIR provides paired visible and infrared images for traffic-oriented perception, with 4,129 pairs for training and 1,013 pairs for testing. Since some visible images contain black borders that may introduce artificial structures during translation, we remove these borders and apply the same spatial preprocessing to both modalities before resizing them to the model input resolution. KAIST contains 10,845 aligned pairs of visible and infrared images captured in urban driving scenarios, including 8,593 pairs for training and 2,252 pairs for testing.

To generate additional infrared training samples for downstream object detection, we use the 2,975 training images and 500 validation images from Cityscapes as an external visible-image pool. All 3,475 images are translated into the infrared style of the corresponding target dataset, and their annotations are converted into detection labels. According to the category definitions of the target benchmarks, we retain \emph{person}, \emph{car}, \emph{motorcycle}, \emph{truck}, \emph{traffic light}, and \emph{bus} for M$^{3}$FD, and \emph{person}, \emph{car}, and \emph{bicycle} for FLIR.

\noindent \textbf{Evaluation metrics.}
We evaluate visible-to-infrared translation from both pixel-level fidelity and perceptual quality. For pixel-level comparison, we report Structural Similarity Index (SSIM)~\cite{wang2004image} and Peak Signal-to-Noise Ratio (PSNR) between translated infrared images and ground-truth infrared images. SSIM reflects structural consistency, while PSNR measures reconstruction accuracy in the image space. However, high pixel-level similarity does not always indicate realistic infrared generation, especially when multiple plausible thermal appearances may correspond to the same visible scene. Therefore, we further report Learned Perceptual Image Patch Similarity (LPIPS)~\cite{zhang2018unreasonable} and Fréchet Inception Distance (FID)~\cite{heusel2017gans}, which better capture perceptual similarity and distribution-level realism. For downstream object detection, we adopt standard detection metrics, including mean Average Precision at IoU 0.5 ($m$AP$_{50}$), mean Average Precision at IoU 0.75 ($m$AP$_{75}$), and the COCO-style mean Average Precision averaged over IoU thresholds from 0.5 to 0.95 ($m$AP)~\cite{lin2014microsoft}. Together, these metrics assess translation fidelity, perceptual quality, distribution-level realism, and downstream detection utility.

\noindent \textbf{Implementation Details.} 
Our SC-Diff is implemented in PyTorch and trained on three NVIDIA A6000 GPUs. During training, all images are center-cropped into square patches according to the shorter side and then resized to $512\times 512$. The pixel values are normalized to $[-1, 1]$, following the same preprocessing protocol as PID~\cite{mao2026pid}. We then use the pretrained autoencoder of SD v1.5 to encode visible images, infrared images, and semantic maps into the latent space, and train the diffusion model with 1,000 diffusion steps to predict the timestep noise. For SAM3, we use a fixed set of text prompts covering common urban-scene categories: \emph{road}, \emph{sidewalk}, \emph{person}, \emph{rider}, \emph{car}, \emph{truck}, \emph{bus}, \emph{motorcycle}, \emph{bicycle}, \emph{building}, \emph{wall}, \emph{pole}, \emph{tree}, \emph{traffic sign}, \emph{traffic light}, \emph{ground}, and \emph{sky}. This prompt set is adopted because M$^{3}$FD, FLIR, KAIST, and Cityscapes are all collected from urban street scenes, where these categories frequently appear. During inference, DDIM sampling with 30 steps is adopted to accelerate the generation process. Considering the spatial resolution of intermediate features, the proposed SGSC module is applied only at the $64\times 64$ and $32\times 32$ resolutions, where token-level semantic labels can still provide sufficiently reliable region guidance. More implementation details are reported in Table~\ref{table_0}. When reproducing competing methods, we use the official implementations and recommended hyperparameters whenever available.
\subsection{Main Results on Visible-to-Infrared Image Translation}
To evaluate the effectiveness of SC-Diff, we compare it with representative visible-to-infrared translation methods from both GAN-based and diffusion-based families. The GAN-based methods include Pix2Pix~\cite{isola2017image}, CycleGAN~\cite{zhu2017unpaired}, ThermalGAN~\cite{kniaz2018thermalgan}, InfraGAN~\cite{ozkanouglu2022infragan}, EGGAN-U~\cite{lee2023edge}, and StegoGAN~\cite{wu2024stegogan}. The diffusion-based methods include T2V-DDPM~\cite{nair2023t2v}, F-ViTA~\cite{paranjape2026f}, DiffV2IR~\cite{ran2025diffv2ir}, and PID~\cite{mao2026pid}. We report results under two evaluation settings. The standard setting uses the train/test split from the same benchmark dataset, while the out-of-distribution setting evaluates whether a trained translator can generalize to external visible images.
\subsubsection{Standard Generation} 
As shown in Table~\ref{table_1}, recent diffusion-based methods generally achieve stronger perceptual performance than most GAN-based baselines. Among all compared methods, SC-Diff achieves the best LPIPS and FID across all three datasets while maintaining competitive PSNR and SSIM. Although PID obtains higher PSNR and SSIM on KAIST, its higher LPIPS and FID indicate weaker perceptual quality and distribution matching. This discrepancy shows that pixel-level metrics alone do not fully capture the visual quality of generated images. Overall, SC-Diff achieves a more favorable balance between perceptual quality and reconstruction fidelity while better preserving local structures.
\begin{table}[t]
\centering
\caption{Out-of-distribution evaluation on Cityscapes under three target infrared styles. FID is computed against the real infrared training distribution of each target dataset, and lower is better.}
\label{tab:ood}
\setlength{\tabcolsep}{9pt}
\renewcommand{\arraystretch}{1.2}
\begin{tabular}{l|c|c|c}
\hline
Method & CS$\rightarrow$M$^3$FD & CS$\rightarrow$FLIR  & CS$\rightarrow$KAIST   \\
\hline
PID~\cite{mao2026pid}  & 191.17 &100.71 &90.76\\ 
F-ViTA~\cite{paranjape2026f} & 156.10 &93.56  &102.33   \\
SC-Diff (Ours)  &\textbf{142.96} &\textbf{89.83} &\textbf{70.28}  \\
\hline
\end{tabular}
\end{table}

\subsubsection{Out-of-Distribution Generation}
We use Cityscapes~\cite{cordts2016cityscapes} (CS) to evaluate out-of-distribution generation. Each translation method is trained separately on the three paired datasets and then used to translate images from Cityscapes without fine-tuning. We evaluate three settings: CS$\rightarrow$M$^{3}$FD, CS$\rightarrow$FLIR, and CS$\rightarrow$KAIST. Under this protocol, we compare SC-Diff with two representative diffusion-based methods, F-ViTA~\cite{paranjape2026f} and PID~\cite{mao2026pid}. This protocol evaluates whether the model can transfer the target infrared appearance to unseen visible scenes, which is important when external visible images are used for data expansion.

Since Cityscapes does not provide paired ground-truth infrared images, we cannot compute metrics that require paired references. We use FID as the only quantitative metric to compare the generated image distribution with the real infrared training distribution of each target dataset. As shown in Table~\ref{tab:ood}, SC-Diff achieves the lowest FID in all three evaluation settings, outperforming F-ViTA and PID. These results indicate that SC-Diff more closely matches the target infrared distributions when translating unseen visible images from Cityscapes.

\subsection{Visualization of Visible-to-Infrared Image Translation}
The qualitative comparisons under the standard in-domain setting are shown in Fig.~\ref{fig_3}. Each row presents a representative example from M$^{3}$FD, FLIR, or KAIST, with the highlighted regions focusing on a vehicle, a cyclist, and small foreground targets, respectively. Most compared methods preserve the overall scene layout, but clear differences emerge in local structure preservation. StegoGAN retains coarse scene content but produces uneven contrast and weakened object details. DiffV2IR and F-ViTA generate plausible infrared appearances, although their local intensity distributions and target contours sometimes differ from the paired infrared images. PID exhibits the most noticeable over-smoothing. The vehicle boundary becomes blurred, while the cyclist and pedestrian structures are substantially weakened. In contrast, SC-Diff preserves more complete target structures and clearer boundaries while maintaining coherent local infrared contrast. Overall, these comparisons show that SC-Diff achieves a better balance between infrared appearance synthesis and local structure preservation.

The qualitative results for out-of-distribution generation are shown in Fig.~\ref{fig_4}. In this setting, visible images from Cityscapes are translated into different target infrared styles without fine-tuning or paired infrared references. This setting is more challenging because the scene distribution differs from the paired training datasets. Existing diffusion-based approaches can generate infrared-like images, but they may introduce inconsistent textures or weaken salient targets when the input scene contains unfamiliar layouts. By comparison, SC-Diff produces more stable translations across the three target styles. The generated images preserve the semantic layout of the visible scene while maintaining clearer object boundaries and more coherent thermal contrast. The zoomed regions further show that SC-Diff preserves finer object details and more coherent local contrast under distribution shift. This property is important when the translated Cityscapes images are used for downstream data augmentation. These observations indicate that SC-Diff transfers the target infrared appearance to unseen Cityscapes images while preserving scene content.

\begin{table*}[htbp]
\centering
\caption{Infrared object detection results on M$^3$FD and FLIR after augmenting the training sets. For each target dataset, visible images from Cityscapes are translated into the corresponding infrared style and added to the original training set. The best results are shown in bold, and the second-best are \underline{underlined}.}
\label{ds_od}
\renewcommand{\arraystretch}{1.2}
\setlength{\tabcolsep}{4pt}

\resizebox{\textwidth}{!}{
\begin{tabular}{l|ccc|ccc|ccc|ccc|ccc}
\hline
\multirow{2}{*}{M$^3$FD} 
& \multicolumn{3}{c|}{RT-DETR} 
& \multicolumn{3}{c|}{YOLOv8} 
& \multicolumn{3}{c|}{FCOS} 
& \multicolumn{3}{c|}{Faster R-CNN}
& \multicolumn{3}{c}{Overall} \\
\cline{2-16} 
& $mAP_{50}$ & $mAP_{75}$ & $mAP$
& $mAP_{50}$ & $mAP_{75}$ & $mAP$
& $mAP_{50}$ & $mAP_{75}$ & $mAP$
& $mAP_{50}$ & $mAP_{75}$ & $mAP$ 
& $mAP_{50}$ & $mAP_{75}$ & $mAP$ \\
\hline
Vanilla  
&47.2  &26.2  &27.0 
&48.6  &28.9  &28.7 
&30.5  &15.8  &15.9 
&35.2  &14.4  &16.9
&40.4  &21.3  &22.1\\
PID~\cite{mao2026pid}      
&46.3  &27.6  &27.6  
&49.7  &\underline{31.4}  &30.6  
&29.4  &15.3  &15.9
&35.8  &16.9  &18.5
&40.3  &22.8  &23.2\\
F-ViTA~\cite{paranjape2026f}   
&\underline{50.6}  &\textbf{30.1} &\underline{30.1}      
&\underline{51.8}  &31.2  &\underline{31.1}  
&\underline{32.8}  &\underline{16.9}  &\underline{17.7}
&\underline{37.0}  &\underline{17.4} &\underline{19.2}
	&\underline{43.1} &\underline{23.9} &\underline{24.5}\\
SC-Diff   
&\textbf{52.5} &\underline{29.8} &\textbf{30.7} 
&\textbf{53.4} &\textbf{34.1} &\textbf{33.5} 
&\textbf{34.3} &\textbf{18.7} &\textbf{18.7}
&\textbf{39.6} &\textbf{18.9} &\textbf{20.9} 
&\textbf{45.0} &\textbf{25.4} &\textbf{26.0}\\
\hline
\end{tabular}
}

\resizebox{\textwidth}{!}{
\begin{tabular}{l|ccc|ccc|ccc|ccc|ccc}
\hline
\multirow{2}{*}{FLIR} 
& \multicolumn{3}{c|}{RT-DETR} 
& \multicolumn{3}{c|}{YOLOv8} 
& \multicolumn{3}{c|}{FCOS} 
& \multicolumn{3}{c|}{Faster R-CNN}
& \multicolumn{3}{c}{Overall} \\
\cline{2-16} 
& $mAP_{50}$ & $mAP_{75}$ & $mAP$
& $mAP_{50}$ & $mAP_{75}$ & $mAP$
& $mAP_{50}$ & $mAP_{75}$ & $mAP$
& $mAP_{50}$ & $mAP_{75}$ & $mAP$ 
& $mAP_{50}$ & $mAP_{75}$ & $mAP$ \\
\hline
Vanilla  
&77.2 &35.9 &40.2  
&78.0 &36.8 &41.2  
&64.0 &19.2 &27.5
	&71.5 &29.3 &\underline{34.9} 
&72.7 &30.3 &36.0\\
PID~\cite{mao2026pid}      
	&77.5  &36.8 &\underline{41.2} 
&79.3  &37.7 &42.0
&62.2  &19.6  &27.5
	&\underline{72.9} &\underline{29.6} &\underline{34.9}
&73.0 &30.9 &36.4\\
F-ViTA~\cite{paranjape2026f}    
	&\underline{78.1}  &\underline{36.9}  &41.1
&\underline{79.7}  &\textbf{38.7}  &\underline{42.3}
&\textbf{66.1} &\underline{20.9} &\underline{29.0}
	&72.3 &28.0  &34.2
&\underline{74.1} &\underline{31.1} &\underline{36.7}\\
SC-Diff   
&\textbf{78.6} &\textbf{38.2}  &\textbf{41.8}
&\textbf{80.2} &\underline{38.6} &\textbf{42.9} 
&\underline{65.6} &\textbf{22.3} &\textbf{29.7} 
&\textbf{73.8} &\textbf{29.7} &\textbf{35.9} 
&\textbf{74.6} &\textbf{32.2} &\textbf{37.6}\\
\hline
\end{tabular}
}
\end{table*}

\begin{figure*}[!t]
	\centering
	\includegraphics[width=\textwidth]{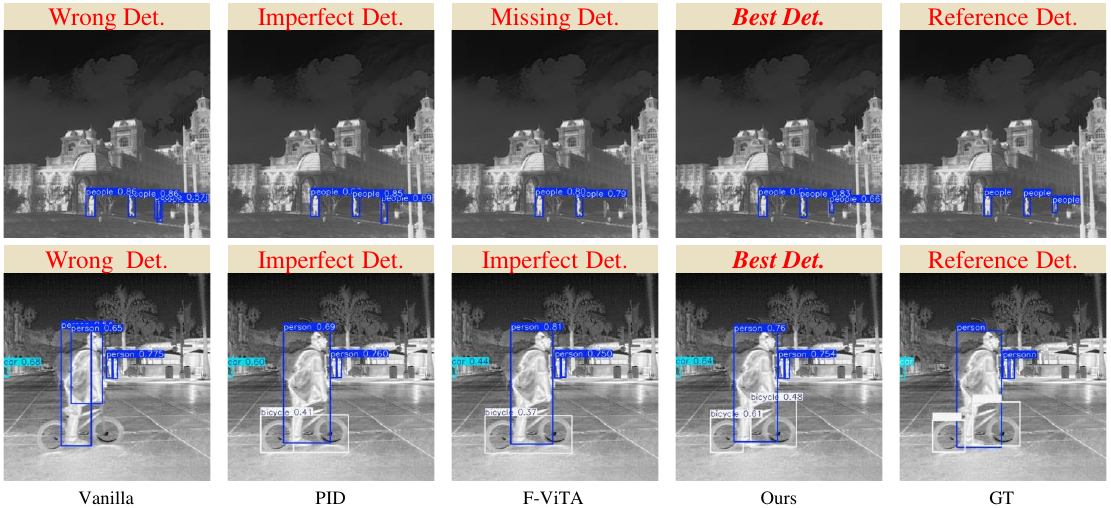}
\caption{Qualitative comparison of YOLOv8 detection results on M$^3$FD and FLIR under different data augmentation settings. The last column shows the ground-truth annotations, and SC-Diff leads to more complete and accurate detections than the compared augmentation methods.}
	\label{fig_5}
\end{figure*}
\subsection{Data Augmentation Performance Evaluation}
\subsubsection{Quantitative Comparison}
To further examine whether the translated infrared images are useful beyond qualitative evaluation, we assess their effectiveness for downstream infrared object detection. Specifically, we use Cityscapes as an external visible-image source and translate its images into the infrared styles of M$^3$FD and FLIR. The translated images are then added to the original training sets, while the test sets remain unchanged. This setting evaluates whether a translation method can generate augmented samples that provide useful supervision rather than merely producing perceptually plausible infrared images. Four representative detectors, including RT-DETR~\cite{zhao2024detrs}, YOLOv8~\cite{Jocher_Ultralytics_YOLO_2023}, FCOS~\cite{tian2019fcos}, and Faster R-CNN~\cite{ren2016faster}, are adopted to provide a comprehensive evaluation across different detection architectures.
The quantitative results are reported in Table~\ref{ds_od}. Compared with the vanilla training setting, augmenting the training set with translated infrared images improves detection performance in most cases, confirming the practical value of visible-to-infrared translation for data expansion. Among the compared augmentation methods, SC-Diff achieves the highest overall $m$AP$_{50}$, $m$AP$_{75}$, and $m$AP on both M$^3$FD and FLIR. On M$^3$FD, it raises the overall $m$AP from 22.1 to 26.0, a gain of 3.9 points. On FLIR, it raises the overall $m$AP from 36.0 to 37.6, a gain of 1.6 points. These consistent gains across multiple detectors suggest that the generated images provide useful information about target objects for detector training.

The comparison with PID and F-ViTA further shows that perceptual realism alone is not sufficient for effective data augmentation. PID introduces physics-based constraints to improve thermal plausibility, but its translated images can be overly smooth, which may weaken object details that detectors rely on. F-ViTA benefits from vision foundation model guidance, yet it mainly uses priors as external guidance and does not explicitly regulate internal token interaction during denoising. As a result, target structures may still be weakened in challenging scenes. These results suggest that SC-Diff benefits detector training because semantic priors are used to organize feature interactions inside the generator, leading to clearer object contours and more distinguishable target regions.
\subsubsection{Qualitative Results} 
The qualitative detection results in Fig.~\ref{fig_5} provide further evidence of the effectiveness of SC-Diff for data augmentation. We select representative samples from M$^3$FD and FLIR and compare the detection outputs obtained after training with different translated infrared images. Detectors trained with SC-Diff augmentation produce more complete detections and more accurate localization in challenging regions. In contrast, detectors trained with PID or F-ViTA augmentation are more likely to miss difficult objects or generate incomplete bounding boxes. These differences are consistent with the quantitative results in Table~\ref{ds_od}. By better preserving object contours and local structures while maintaining target saliency during translation, SC-Diff may provide more informative supervision for detector training.
\subsection{Ablation Study}
\label{experiments_d}
\subsubsection{Effect of Clean Condition Encoding and SGSC Components}
\begin{figure}[t]
\centering
\includegraphics[width=1.0\columnwidth]{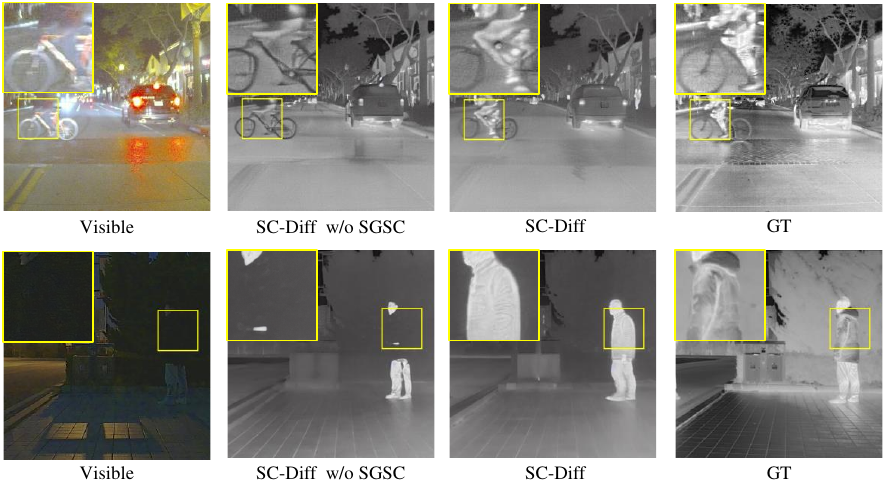}
\caption{Qualitative comparison of SGSC. Yellow boxes mark the enlarged regions for comparison.}
\label{fig_6}
\end{figure}

Table~\ref{tab:ablation_sgsc} evaluates CCE and the two SGSC components, $\beta$ and $G$. CCE uses only the visible image and semantic map as clean conditions, without introducing the noisy infrared latent. This reduces FID from 65.62 to 63.35. With CCE fixed, $\beta$ assigns the same bias to all same-category keys for each query. The resulting improvement is limited, indicating that query-wise scalar calibration alone is insufficient. $G$ uses the original attention weights to assign larger bias increments to same-category keys with stronger responses. This modulation focuses the calibration on more relevant interactions and reduces FID to 57.66, while LPIPS remains comparable. As shown in Fig.~\ref{fig_6}, the full SGSC also produces clearer contours and better-preserved local structures in the highlighted regions.

\begin{table}[!t]
\centering
\caption{Ablation of clean condition encoding (CCE) and SGSC components on FLIR. The best result for each metric is shown in \textbf{bold}.}
\label{tab:ablation_sgsc}
\vspace{-4pt}
\setlength{\tabcolsep}{3.0pt}
\renewcommand{\arraystretch}{1.12}
\resizebox{\columnwidth}{!}{%
\begin{tabular}{c@{\hspace{14pt}}c@{\hspace{14pt}}cccc}
\hline
\multirow{2}{*}{CCE}
& SGSC
& \multicolumn{4}{c}{Metrics} \\
\cmidrule(l{5pt}r{15pt}){2-2}\cmidrule(lr){3-6}
& \makebox[1.08cm][c]{\makebox[0.54cm][c]{$\beta$}\makebox[0.54cm][c]{$G$}}
& PSNR ($\uparrow$) 
& SSIM ($\uparrow$) 
& LPIPS ($\downarrow$) 
& FID ($\downarrow$) \\
\hline
 & \makebox[1.08cm][c]{\makebox[0.54cm][c]{}\makebox[0.54cm][c]{}} &18.32 &0.480  &0.267 &65.62 \\
\checkmark & \makebox[1.08cm][c]{\makebox[0.54cm][c]{}\makebox[0.54cm][c]{}} &18.70 &0.485  &\textbf{0.262} &63.35 \\
\checkmark & \makebox[1.08cm][c]{\makebox[0.54cm][c]{\checkmark}\makebox[0.54cm][c]{}} &18.68   &0.483   &0.270   &61.74   \\
\checkmark & \makebox[1.08cm][c]{\makebox[0.54cm][c]{\checkmark}\makebox[0.54cm][c]{\checkmark}} &\textbf{19.15}   &\textbf{0.496}   &\textbf{0.262}   &\textbf{57.66}   \\
\hline
\end{tabular}
}
\end{table}
\subsubsection{Influence of $c_s$ and $c_v$}
\begin{table}[t]
\centering
\caption{Effect of CFG scales on FLIR. The best results are shown in \textbf{bold}, and the second-best are \underline{underlined}.}
\label{tab:cfg_scale}
\setlength{\tabcolsep}{8pt}
\renewcommand{\arraystretch}{1.2}
\begin{tabular}{cc|cccc}
\toprule
\multicolumn{2}{c|}{CFG Scale} 
& \multirow{2}{*}{PSNR($\uparrow$)} 
& \multirow{2}{*}{SSIM($\uparrow$)} 
& \multirow{2}{*}{LPIPS($\downarrow$)} 
& \multirow{2}{*}{FID($\downarrow$)} \\
\cmidrule(r){1-2}
$c_s$ & $c_v$ &  &  &  &  \\
\midrule
\multirow{3}{*}{0.5}
& 0.5 &16.00  &0.435  &0.410   &127.63  \\
& 1.5 &18.89  &\textbf{0.496}  &0.264  &\textbf{56.80} \\
& 2.0 &\underline{18.96}  &\underline{0.495}  &\underline{0.262}  &61.98  \\
\midrule
\multirow{3}{*}{2.0}
& 0.5 &16.19  &0.437  &0.386   &98.18 \\
& 1.5 &\textbf{19.15}  &\textbf{0.496}  &\underline{0.262}  &\underline{57.66} \\
& 2.0 &18.94  &\underline{0.495}  &\textbf{0.260}  &61.71  \\
\midrule
\multirow{3}{*}{3.5}
& 0.5 &15.48  &0.436  &0.373   &77.54 \\
& 1.5 &18.64  &0.491  &0.268  &59.97 \\
& 2.0 &18.81  &0.493  &0.263  &62.16  \\
\bottomrule
\end{tabular}
\vspace{10pt}
	\caption{Effect of DDIM sampling steps on $\text{M}^3$FD, FLIR, and KAIST. The best results are shown in \textbf{bold}, and the second-best are \underline{underlined}.}
	\label{table_5}
    \renewcommand{\arraystretch}{1.2}
	\resizebox{0.48\textwidth}{!}{%
	\begin{tabular}{c|c|cccc}
	\toprule
	Datasets & Steps & PSNR $(\uparrow)$ &SSIM $(\uparrow)$ &LPIPS $(\downarrow)$ &FID $(\downarrow)$\\ 
	\midrule
	\multirow{4}{*}{$\text{M}^3$FD} & 10 &20.09 &0.718 &0.189 &67.09 \\
	 &30 &\textbf{20.63} & \textbf{0.731} &\textbf{0.181} &62.91  \\
	 &50 &\underline{20.54} &\underline{0.728} &\underline{0.182} &\underline{62.42} \\
	 &100 &\underline{20.54} &0.726 &\underline{0.182} &\textbf{62.23}\\ \hline
	\multirow{4}{*}{FLIR} &10 &18.63 &\textbf{0.505} &0.275 &61.38\\
	 &30 &\textbf{19.15} &\underline{0.496} &\underline{0.262} &57.66\\
	 &50 &\underline{19.02} &0.487 &\underline{0.262} &\underline{56.51} \\
	 &100 &19.01 &0.484 &\textbf{0.260} &\textbf{56.41} \\ \hline
		\multirow{4}{*}{KAIST} &10 &22.07 &0.765 &0.248 &33.77 \\
	 &30 &\textbf{23.49} &\textbf{0.799} &\textbf{0.239} &\textbf{27.98} \\
	 &50 &23.36 &\underline{0.795} &\underline{0.242} &\underline{28.61} \\
	 &100 &\underline{23.45} &\underline{0.795} &0.243 &28.67 \\
		\bottomrule
		\end{tabular}
		}
	\end{table}
\begin{figure}[!t]
	\centering
	\includegraphics[width=\columnwidth]{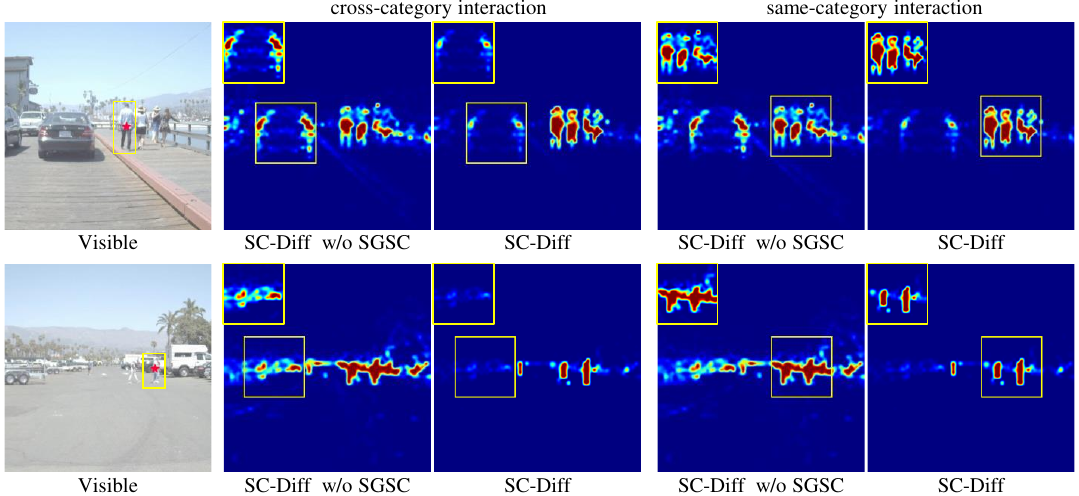}
			\caption{Visualization of self-attention maps from the final U-Net layer. Red stars mark query points, and yellow boxes show enlarged regions. SC-Diff reduces cross-category responses and concentrates more attention on same-category regions.}
	\label{fig_7}
	\vspace{10pt}
	\centering
	\includegraphics[width=\columnwidth]{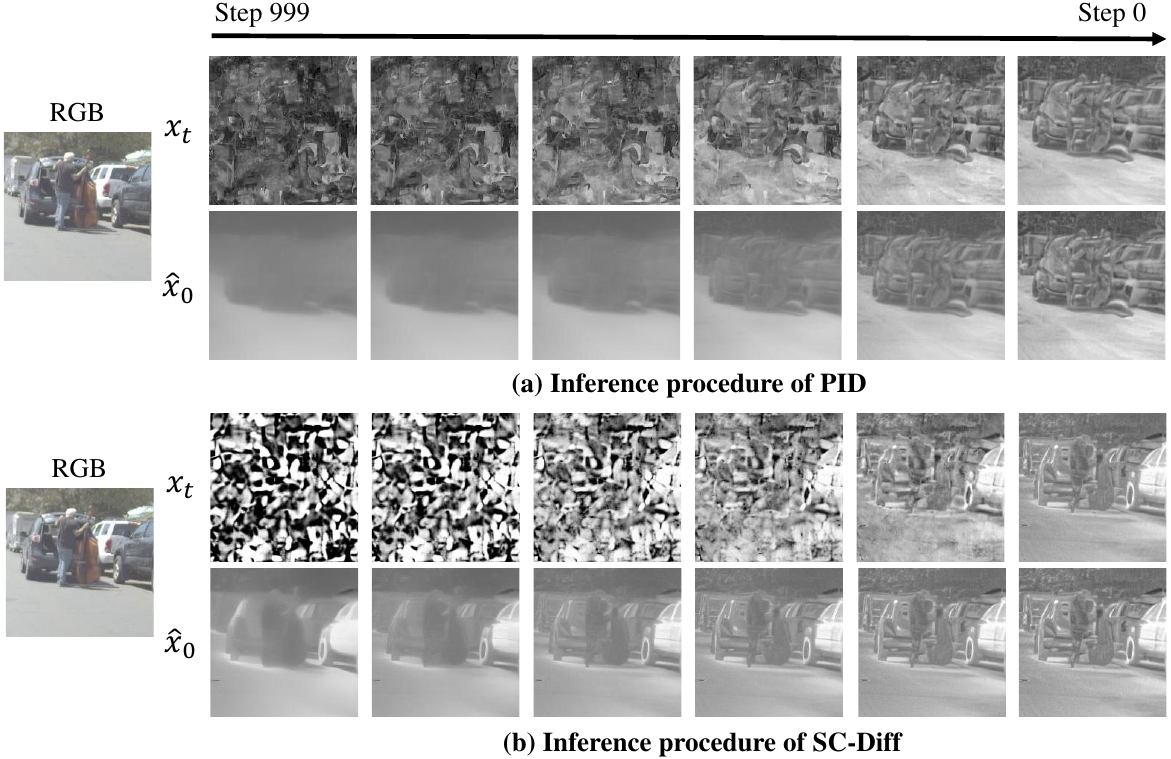}
		\caption{Visualization of intermediate results during inference. Both PID and SC-Diff progressively generate infrared images through iterative refinement.}
	\label{fig_8}
\end{figure}
We study the effects of the semantic guidance scale ($c_s$) and the visible guidance scale ($c_v$) on FLIR, as shown in Table~\ref{tab:cfg_scale}. With $c_s$ fixed at 0.5, setting $c_v$ to 1.5 produces the lowest FID, whereas setting it to 2.0 gives a slightly higher PSNR. Changing $c_v$ from 1.5 to 2.0 has little effect on the reconstruction and perceptual metrics but consistently increases FID. A large $c_s$ may impose excessive constraints on the denoising process and impair infrared texture generation. Based on these results, we use 2.0 and 1.5 as the default values of $c_s$ and $c_v$, respectively, because this setting performs consistently well across all metrics.
\subsubsection{Influence of sampling steps} 
We examine the effect of DDIM sampling steps, as shown in Table~\ref{table_5}. The results show that 30 steps improve most metrics over 10 steps across all three datasets. This comparison indicates that 10 steps do not provide sufficient refinement. Across all three datasets, PSNR decreases slightly when the number of steps exceeds 30. FID continues to decrease on M$^3$FD and FLIR, but increases on KAIST at 50 steps. We select 30 steps as the default setting to balance translation quality and inference cost.
\subsection{Visualization}
\label{ex_vis}
\subsubsection{\textbf{Visualization of Self-Attention Maps}}
\label{ex_attn_vis}
Fig.~\ref{fig_7} visualizes the self-attention maps for selected query points marked by red stars. We compare SC-Diff w/o SGSC and SC-Diff under cross-category and same-category interaction cases. Without SGSC, attention responses may spread to semantically unrelated regions, leading to noticeable cross-category attention leakage. After applying SGSC, cross-category attention is reduced, while attention becomes more concentrated on same-category tokens. Meanwhile, SGSC suppresses rather than eliminates cross-category responses, thereby retaining global contextual interactions. This visualization shows that semantic priors regulate internal token aggregation during denoising rather than only serving as external layout guidance.
\subsubsection{\textbf{Visualization of intermediate results}}
To better understand the generation process of SC-Diff, we visualize the intermediate denoising results in Fig.~\ref{fig_8}. At the early sampling steps, both PID and SC-Diff recover coarse scene layouts from noisy latent representations. As denoising proceeds, SC-Diff first clarifies object boundaries and then refines local structures with more coherent thermal contrast. This process leads to clearer and more complete infrared images. Compared with PID, SC-Diff maintains local details more consistently throughout the sampling trajectory instead of relying on late-stage refinement. This observation suggests that semantic-guided calibration provides stable structural guidance during iterative denoising and helps preserve the correspondence between the visible input and the generated infrared image.
\section{Conclusion}
\label{conclusion}
This paper presented SC-Diff, a semantically calibrated latent diffusion framework for visible-to-infrared image translation. SC-Diff goes beyond conventional external conditioning by introducing semantic priors into the internal self-attention layers of the denoising U-Net. Through the proposed SGSC module, same-category token interactions are adaptively enhanced when the original attention distribution is semantically dispersed, which reduces cross-category interference while maintaining soft global contextual dependencies. Experiments on multiple public datasets show that SC-Diff improves perceptual generation quality and provides effective augmented data for infrared object detection.
\IEEEtriggeratref{39}
\IEEEtriggercmd{\enlargethispage{2\baselineskip}}
\bibliographystyle{IEEEtran_yearlast}
\bibliography{refs} 

\begin{thebibliography}{10}
\providecommand{\url}[1]{#1}
\csname url@samestyle\endcsname
\providecommand{\newblock}{\relax}
\providecommand{\bibinfo}[2]{#2}
\providecommand{\BIBentrySTDinterwordspacing}{\spaceskip=0pt\relax}
\providecommand{\BIBentryALTinterwordstretchfactor}{4}
\providecommand{\BIBentryALTinterwordspacing}{\spaceskip=\fontdimen2\font plus
\BIBentryALTinterwordstretchfactor\fontdimen3\font minus
  \fontdimen4\font\relax}
\providecommand{\BIBforeignlanguage}[2]{{%
\expandafter\ifx\csname l@#1\endcsname\relax
\typeout{** WARNING: IEEEtran.bst: No hyphenation pattern has been}%
\typeout{** loaded for the language `#1'. Using the pattern for}%
\typeout{** the default language instead.}%
\else
\language=\csname l@#1\endcsname
\fi
#2}}
\providecommand{\BIBdecl}{\relax}
\BIBdecl

\bibitem{mao2026pid}
F.~Mao, J.~Mei, S.~Lu, F.~Liu, L.~Chen, F.~Zhao, and Y.~Hu, ``{PID}:
  Physics-informed diffusion model for infrared image generation,''
  \emph{Pattern Recognition}, vol. 169, pp. 111816, 2026.

\bibitem{FLIR_dataset}
{FLIR Team}, ``Free flir thermal dataset for algorithm training,''
  \url{https://www.flir.com/oem/adas/adas-dataset-form/}.

\bibitem{wang2024ov}
H.~Wang, P.~Ren, Z.~Jie, X.~Dong, C.~Feng, Y.~Qian, L.~Ma, D.~Jiang, Y.~Wang,
  X.~Lan \emph{et~al.}, ``{OV-DINO}: Unified open-vocabulary detection with
  language-aware selective fusion,'' \emph{arXiv preprint arXiv:2407.07844},
  2024.

\bibitem{li2023multiscale}
R.~Li, J.~Xiang, F.~Sun, Y.~Yuan, L.~Yuan, and S.~Gou, ``Multiscale cross-modal
  homogeneity enhancement and confidence-aware fusion for multispectral
  pedestrian detection,'' \emph{IEEE Transactions on Multimedia}, vol.~26, pp.
  852--863, 2023.

\bibitem{2024MMI-Det}
Y.~Zeng, T.~Liang, Y.~Jin, and Y.~Li, ``Mmi-det: Exploring multi-modal
  integration for visible and infrared object detection,'' \emph{IEEE
  Transactions on Circuits and Systems for Video Technology}, vol.~34, no.~11,
  pp. 11\,198--11\,213, 2024.

\bibitem{EIDet}
K.~Hu, Y.~He, Y.~Li, J.~Zhao, S.~Chen, and Y.~Kang, ``Ei²det: Edge-guided
  illumination-aware interactive learning for visible-infrared object
  detection,'' \emph{IEEE Transactions on Circuits and Systems for Video
  Technology}, vol.~35, no.~7, pp. 7101--7115, 2025.

\bibitem{2024DPD}
G.~Sun, Z.~Xiong, and Y.~Yuan, ``Detail-preserving and diverse image
  translation for adverse visual object detection,'' \emph{IEEE Transactions on
  Circuits and Systems for Video Technology}, vol.~34, no.~10, pp. 9139--9152,
  2024.

\bibitem{carion2025sam3}
N.~Carion, L.~Gustafson, Y.-T. Hu, S.~Debnath, R.~Hu, D.~Suris, C.~Ryali, K.~V.
  Alwala, H.~Khedr, A.~Huang \emph{et~al.}, ``{SAM3}: Segment anything with
  concepts,'' in \emph{Proceedings of the International Conference on Learning
  Representations}, 2026.

\bibitem{isola2017image}
P.~Isola, J.-Y. Zhu, T.~Zhou, and A.~A. Efros, ``Image-to-image translation
  with conditional adversarial networks,'' in \emph{Proceedings of the IEEE
  Conference on Computer Vision and Pattern Recognition}, pp. 1125--1134, 2017.

\bibitem{zhu2017unpaired}
J.-Y. Zhu, T.~Park, P.~Isola, and A.~A. Efros, ``Unpaired image-to-image
  translation using cycle-consistent adversarial networks,'' in
  \emph{Proceedings of the IEEE International Conference on Computer Vision},
  pp. 2223--2232, 2017.

\bibitem{wu2024stegogan}
S.~Wu, Y.~Chen, S.~Mermet, L.~Hurni, K.~Schindler, N.~Gonthier, and
  L.~Landrieu, ``{StegoGAN}: Leveraging steganography for non-bijective
  image-to-image translation,'' in \emph{Proceedings of the IEEE Conference on
  Computer Vision and Pattern Recognition}, pp. 7922--7931, 2024.

\bibitem{PanopticGAN}
L.~Zhang, P.~Ratsamee, Z.~Luo, Y.~Uranishi, M.~Higashida, and H.~Takemura,
  ``Panoptic-level image-to-image translation for object recognition and visual
  odometry enhancement,'' \emph{IEEE Transactions on Circuits and Systems for
  Video Technology}, vol.~34, no.~2, pp. 938--954, 2024.

\bibitem{ho2020denoising}
J.~Ho, A.~Jain, and P.~Abbeel, ``Denoising diffusion probabilistic models,''
  \emph{Advances in Neural Information Processing Systems}, pp. 6840--6851,
  2020.

\bibitem{rombach2022high}
R.~Rombach, A.~Blattmann, D.~Lorenz, P.~Esser, and B.~Ommer, ``High-resolution
  image synthesis with latent diffusion models,'' in \emph{Proceedings of the
  IEEE Conference on Computer Vision and Pattern Recognition}, pp.
  10\,684--10\,695, 2022.

\bibitem{zhang2023adding}
L.~Zhang, A.~Rao, and M.~Agrawala, ``Adding conditional control to
  text-to-image diffusion models,'' in \emph{Proceedings of the IEEE
  International Conference on Computer Vision}, pp. 3836--3847, 2023.

\bibitem{li2023bbdm}
B.~Li, K.~Xue, B.~Liu, and Y.-K. Lai, ``{BBDM}: Image-to-image translation with
  brownian bridge diffusion models,'' in \emph{Proceedings of the IEEE
  Conference on Computer Vision and Pattern Recognition}, pp. 1952--1961, 2023.

\bibitem{xia2024diffusion}
M.~Xia, Y.~Zhou, R.~Yi, Y.-J. Liu, and W.~Wang, ``A diffusion model translator
  for efficient image-to-image translation,'' \emph{IEEE Trans. Pattern Anal.
  Mach. Intell.}, vol.~46, no.~12, pp. 10\,272--10\,283, 2024.

\bibitem{AdaNoise}
X.~Yang, H.~Shi, F.~Gao, and N.~Wang, ``Adanoise: Cycle-consistent image
  translation with domain-adaptive noise perturbation,'' \emph{IEEE
  Transactions on Circuits and Systems for Video Technology}, pp. 1--1, 2026.

\bibitem{kniaz2018thermalgan}
V.~V. Kniaz, V.~A. Knyaz, J.~Hladuvka, W.~G. Kropatsch, and V.~Mizginov,
  ``{ThermalGAN}: Multimodal color-to-thermal image translation for person
  re-identification in multispectral dataset,'' in \emph{Proceedings of the
  European Conference on Computer Vision Workshops}, 2018.

\bibitem{ozkanouglu2022infragan}
M.~A. {\"O}zkano{\u{g}}lu and S.~Ozer, ``{InfraGAN}: A {GAN} architecture to
  transfer visible images to infrared domain,'' \emph{Pattern Recognition
  Letters}, pp. 69--76, 2022.

\bibitem{lee2023edge}
D.~G. Lee, M.~H. Jeon, Y.~Cho, and A.~Kim, ``Edge-guided multi-domain
  {RGB}-to-{TIR} image translation for training vision tasks with challenging
  labels,'' in \emph{Proceedings of the IEEE International Conference on
  Robotics and Automation}, pp. 8291--8298, 2023.

\bibitem{paranjape2026f}
J.~N. Paranjape, C.~M. De~Melo, and V.~M. Patel, ``F-vita: Foundation model
  guided visible to infrared translation,'' in \emph{Proceedings of the IEEE
  Winter Conference on Applications of Computer Vision}, pp. 5633--5642, 2026.

\bibitem{liu2024grounding}
S.~Liu, Z.~Zeng, T.~Ren, F.~Li, H.~Zhang, J.~Yang, Q.~Jiang, C.~Li, J.~Yang,
  H.~Su \emph{et~al.}, ``Grounding {DINO}: Marrying {DINO} with grounded
  pre-training for open-set object detection,'' in \emph{Proceedings of the
  European Conference on Computer Vision}, pp. 38--55, 2024.

\bibitem{kirillov2023segment}
A.~Kirillov, E.~Mintun, N.~Ravi, H.~Mao, C.~Rolland, L.~Gustafson, T.~Xiao,
  S.~Whitehead, A.~C. Berg, W.-Y. Lo \emph{et~al.}, ``Segment anything,'' in
  \emph{Proceedings of the IEEE International Conference on Computer Vision},
  pp. 4015--4026, 2023.

\bibitem{ran2025diffv2ir}
L.~Ran, L.~Wang, G.~Wang, P.~Wang, and Y.~Zhang, ``{DiffV2IR}:
  Visible-to-infrared diffusion model via vision-language understanding,''
  \emph{arXiv preprint arXiv:2503.19012}, 2025.

\bibitem{lee2026thera}
D.-G. Lee, T.~H. Rhee, H.~Jang, Y.-S. Shin, U.~Shin, and A.~Kim, ``Thera:
  Thermal-aware visual-language prompting for controllable rgb-to-thermal
  infrared translation,'' in \emph{Proceedings of the IEEE Conference on
  Computer Vision and Pattern Recognition}, 2026.

\bibitem{xiao2025thermalgen}
J.~Xiao, R.~Nayak, N.~Zhang, D.~Tortei, and G.~Loianno, ``{ThermalGen}:
  Style-disentangled flow-based generative models for {RGB}-to-thermal image
  translation,'' \emph{Advances in Neural Information Processing Systems},
  vol.~38, 2025.

\bibitem{song2020denoising}
J.~Song, C.~Meng, and S.~Ermon, ``Denoising diffusion implicit models,''
  \emph{arXiv preprint arXiv:2010.02502}, 2020.

\bibitem{ho2022classifier}
J.~Ho and T.~Salimans, ``Classifier-free diffusion guidance,'' \emph{arXiv
  preprint arXiv:2207.12598}, 2022.

\bibitem{nair2023t2v}
N.~G. Nair and V.~M. Patel, ``{T2V-DDPM}: Thermal to visible face translation
  using denoising diffusion probabilistic models,'' in \emph{Proceedings of the
  IEEE International Conference on Automatic Face and Gesture Recognition}, pp.
  1--7, 2023.

\bibitem{liu2022target}
J.~Liu, X.~Fan, Z.~Huang, G.~Wu, R.~Liu, W.~Zhong, and Z.~Luo, ``Target-aware
  dual adversarial learning and a multi-scenario multi-modality benchmark to
  fuse infrared and visible for object detection,'' in \emph{Proceedings of the
  IEEE Conference on Computer Vision and Pattern Recognition}, pp. 5802--5811,
  2022.

\bibitem{HwangKAIST}
S.~Hwang, J.~Park, N.~Kim, Y.~Choi, and I.~S. Kweon, ``Multispectral pedestrian
  detection: Benchmark dataset and baseline,'' in \emph{Proceedings of the IEEE
  Conference on Computer Vision and Pattern Recognition}, pp. 1037--1045, 2015.

\bibitem{cordts2016cityscapes}
M.~Cordts, M.~Omran, S.~Ramos, T.~Rehfeld, M.~Enzweiler, R.~Benenson,
  U.~Franke, S.~Roth, and B.~Schiele, ``The cityscapes dataset for semantic
  urban scene understanding,'' in \emph{Proceedings of the IEEE Conference on
  Computer Vision and Pattern Recognition}, pp. 3213--3223, 2016.

\bibitem{wang2004image}
Z.~Wang, A.~C. Bovik, H.~R. Sheikh, and E.~P. Simoncelli, ``Image quality
  assessment: From error visibility to structural similarity,'' \emph{IEEE
  Transactions on Image Processing}, pp. 600--612, 2004.

\bibitem{zhang2018unreasonable}
R.~Zhang, P.~Isola, A.~A. Efros, E.~Shechtman, and O.~Wang, ``The unreasonable
  effectiveness of deep features as a perceptual metric,'' in \emph{Proceedings
  of the IEEE Conference on Computer Vision and Pattern Recognition}, pp.
  586--595, 2018.

\bibitem{heusel2017gans}
M.~Heusel, H.~Ramsauer, T.~Unterthiner, B.~Nessler, and S.~Hochreiter, ``{GANs}
  trained by a two time-scale update rule converge to a local nash
  equilibrium,'' \emph{Advances in Neural Information Processing Systems},
  2017.

\bibitem{lin2014microsoft}
T.-Y. Lin, M.~Maire, S.~Belongie, J.~Hays, P.~Perona, D.~Ramanan,
  P.~Doll{\'a}r, and C.~L. Zitnick, ``Microsoft {COCO}: Common objects in
  context,'' in \emph{Proceedings of the European Conference on Computer
  Vision}, pp. 740--755, 2014.

\bibitem{zhao2024detrs}
Y.~Zhao, W.~Lv, S.~Xu, J.~Wei, G.~Wang, Q.~Dang, Y.~Liu, and J.~Chen, ``{DETRs}
  beat {YOLOs} on real-time object detection,'' in \emph{Proceedings of the
  IEEE Conference on Computer Vision and Pattern Recognition}, pp.
  16\,965--16\,974, 2024.

\bibitem{Jocher_Ultralytics_YOLO_2023}
\BIBentryALTinterwordspacing
G.~Jocher, J.~Qiu, and A.~Chaurasia, ``{Ultralytics YOLO}.'' [Online].
  Available: \url{https://github.com/ultralytics/ultralytics} Jan. 2023.
\BIBentrySTDinterwordspacing

\bibitem{tian2019fcos}
Z.~Tian, C.~Shen, H.~Chen, and T.~He, ``{FCOS}: Fully convolutional one-stage
  object detection,'' in \emph{Proceedings of the IEEE International Conference
  on Computer Vision}, pp. 9627--9636, 2019.

\bibitem{ren2016faster}
S.~Ren, K.~He, R.~Girshick, and J.~Sun, ``Faster {R-CNN}: Towards real-time
  object detection with region proposal networks,'' \emph{IEEE Transactions on
  Pattern Analysis and Machine Intelligence}, vol.~39, no.~6, pp. 1137--1149,
  2016.

\end{thebibliography}

\newpage
\begin{IEEEbiography}
	[{\includegraphics[width=1in,height=1.25in,clip]{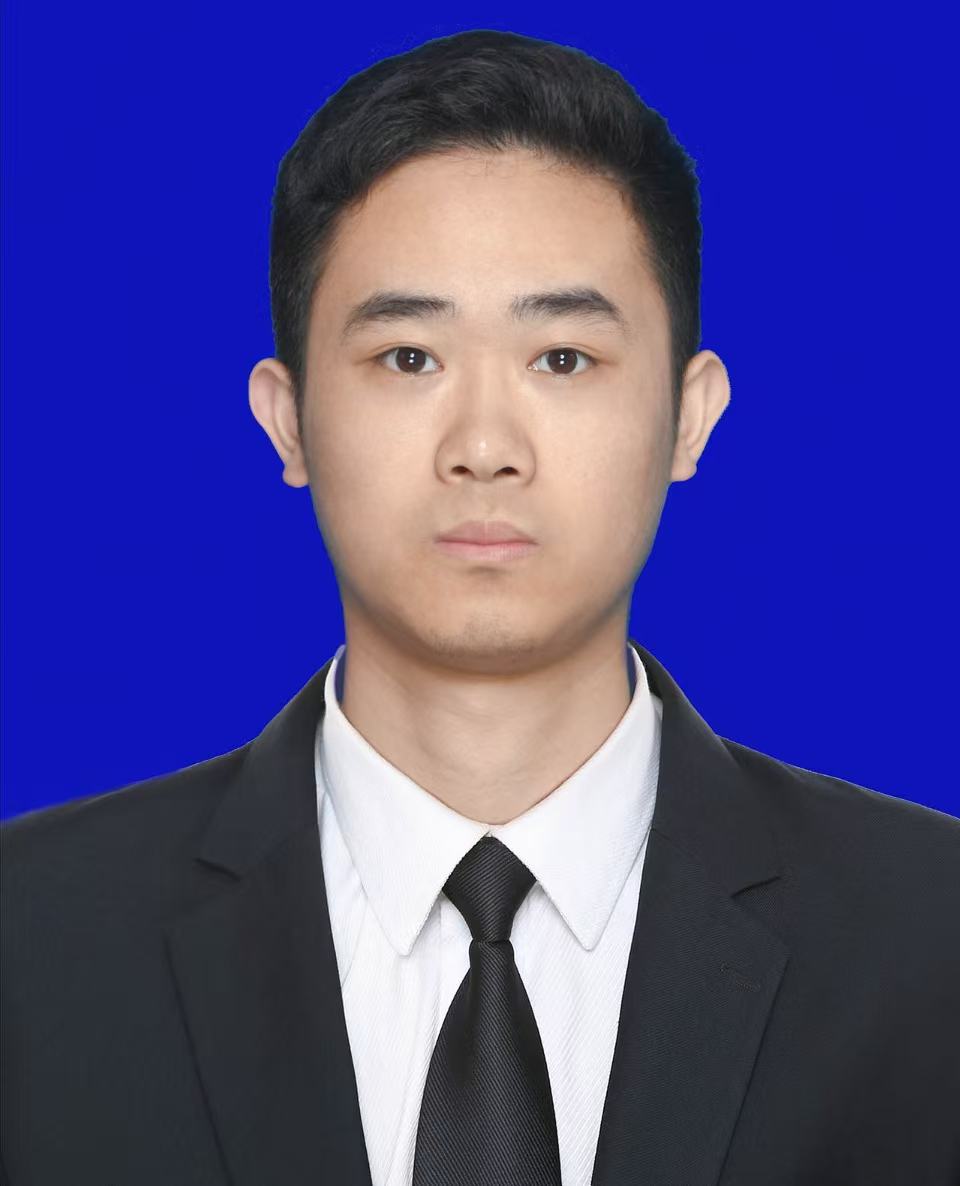}}]
	{Junyin Zhang}
	received the B.S. degree in digital media technology from Chongqing University of Posts and Telecommunications, Chongqing, China, in 2019, and the M.S. degree in software engineering
    from Chongqing University, Chongqing, in 2022. He is currently pursuing the Ph.D. degree in control science and engineering with the School of Intelligent Systems Engineering, Sun Yat-sen University, Shenzhen, China. His research interests include computer vision and image generation.
\end{IEEEbiography}

\begin{IEEEbiography}
	[{\includegraphics[width=1in,height=1.25in,clip]{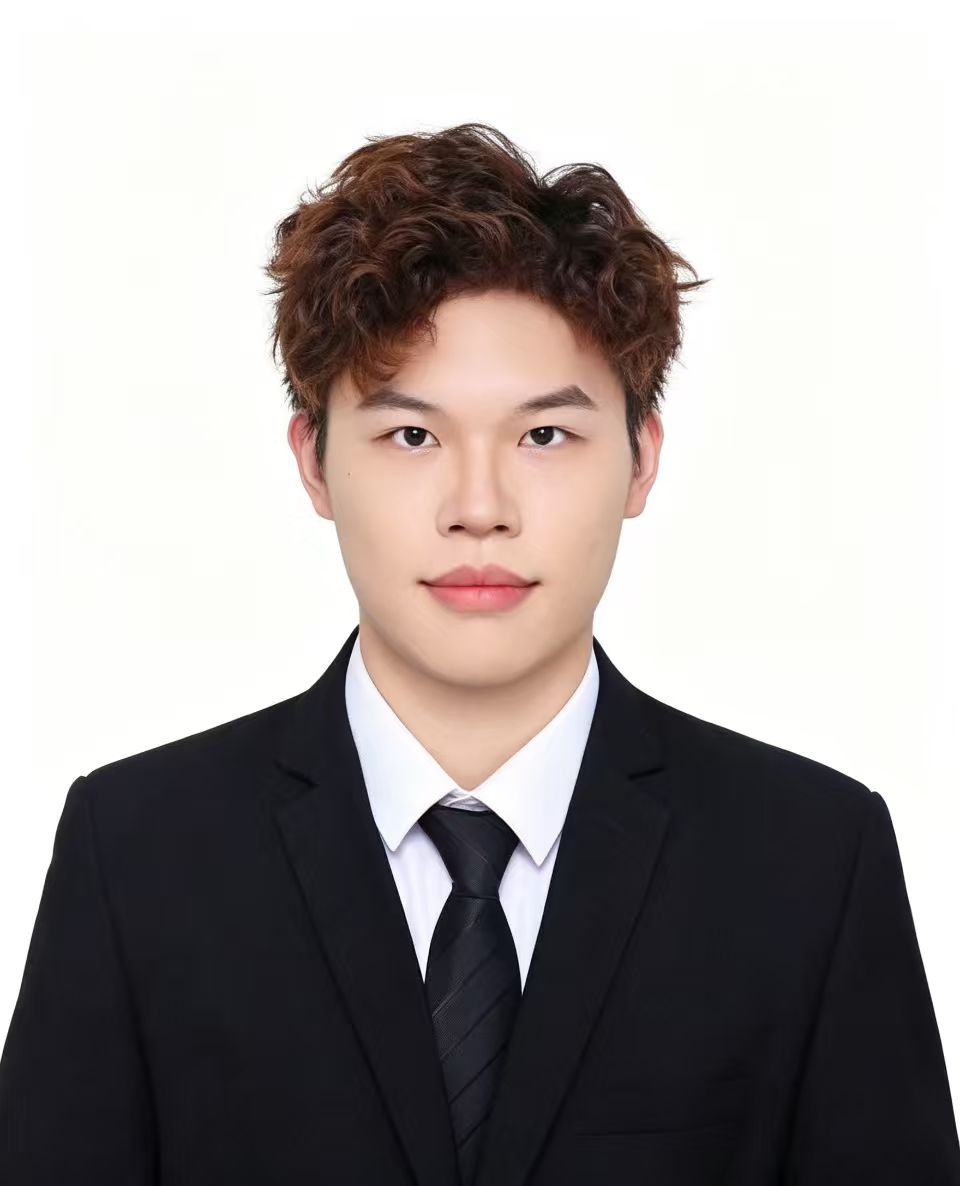}}]
	{Siyu Huang}
	is currently pursuing the B.S. degree in Intelligent Science and Technology from Sun Yat-sen University, Guangdong, China.
	He is particularly interested in long video generation and hope to explore this direction more deeply during his Ph.D. studies. His long-term goal is to develop meaningful and practical AI methods that can contribute to real-world industrial applications.
\end{IEEEbiography}

\begin{IEEEbiography}
	[{\includegraphics[width=1in,height=1.25in,clip]{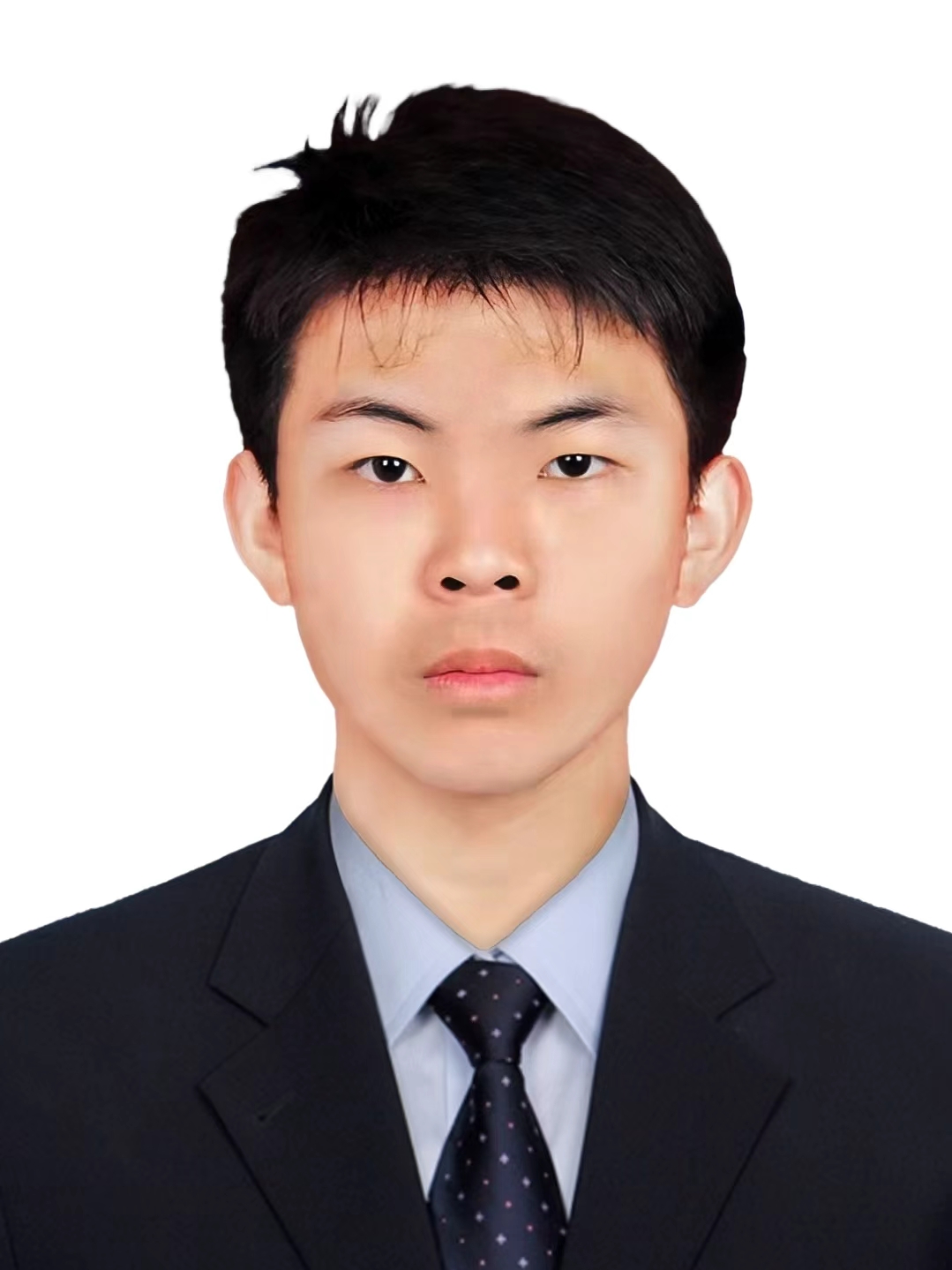}}]
	{Jianxiong Ye} 
	received the B.S. degree in electronic and information engineering from Wuyi University, Jiangmen, China, in 2025. He is currently pursuing the M.S. degree in control science and engineering with the School of Intelligent Systems Engineering, Sun Yat-sen University, Shenzhen, China.
	His research interests include computer vision, pattern recognition, and intelligent robotics. He is working on multi-modal object detection and few-shot learning problems.
\end{IEEEbiography}

\begin{IEEEbiography}
	[{\includegraphics[width=1in,height=1.25in,clip]{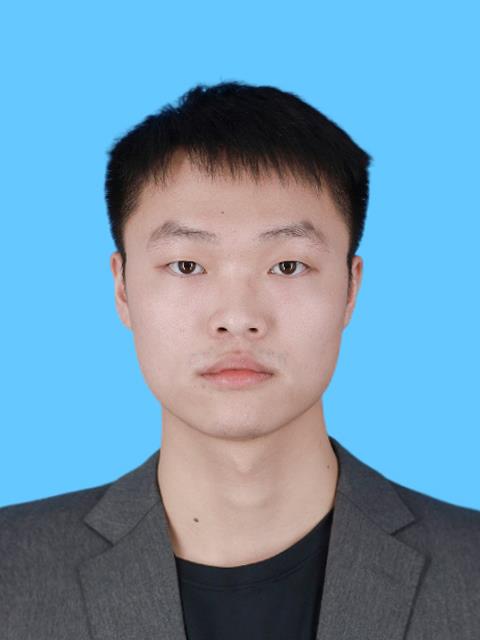}}]
	{Haowei Gong} 
	received the B.S. degree in Intelligent Science and Technology from Sun Yat-sen University, China, in 2026. He is currently pursuing the M.S. degree in Control Science and Engineering with the School of Intelligent Engineering, Sun Yat-sen University, Shenzhen, China. His research interests include computer vision and image generation.
\end{IEEEbiography}

\begin{IEEEbiography}
	[{\includegraphics[width=1in,height=1.25in,clip]{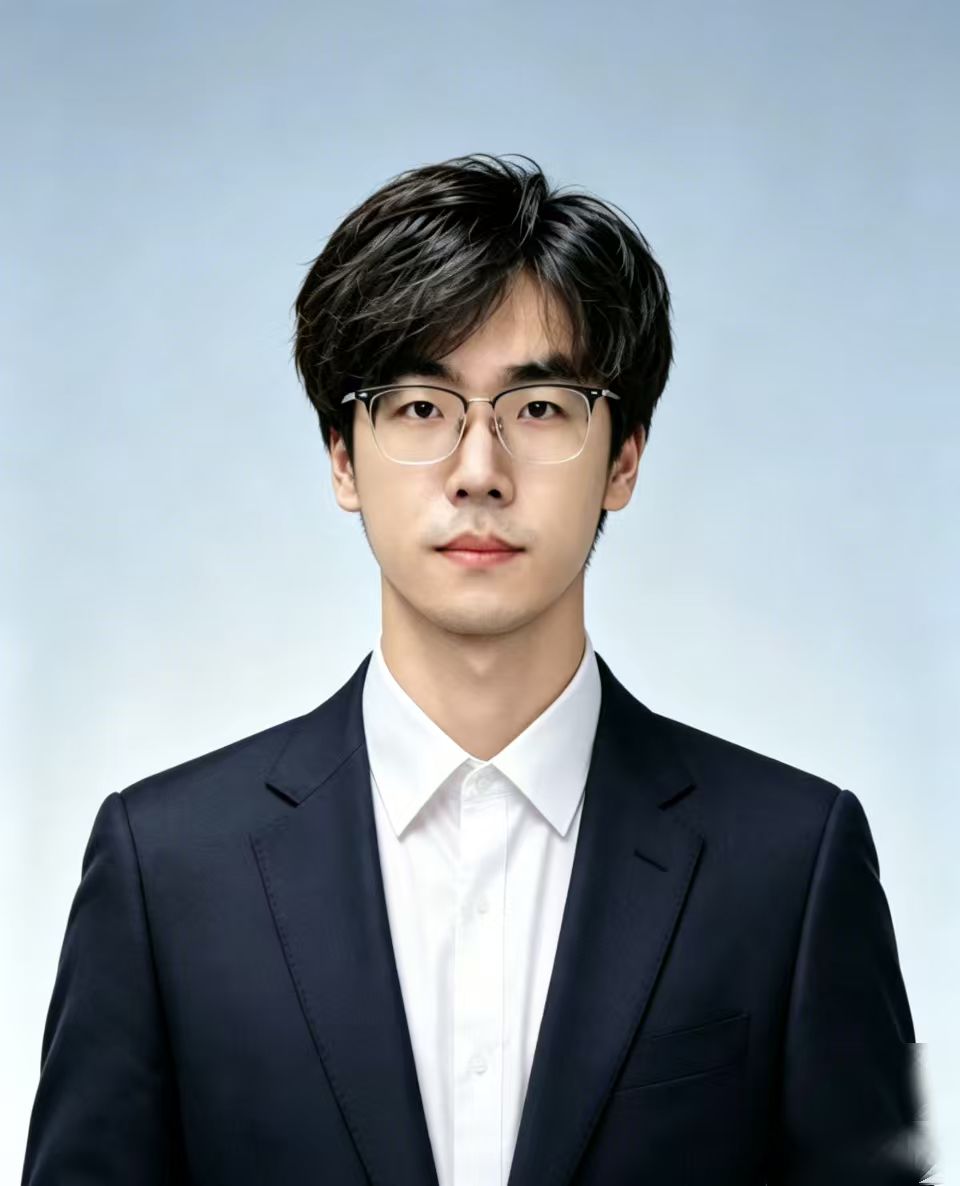}}]
	{Ruicheng Zhang}
	received the B.S. degree in Intelligent Science and Technology from Sun Yat-sen University, Guangdong, China. He is currently pursuing the M.S. degree in Artificial Intelligence at Tsinghua University, Guangdong, China.
	He interned at Video Rebirth and Tencent Project Up. His research interests include video generation and world models.
\end{IEEEbiography}

\begin{IEEEbiography}
	[{\includegraphics[width=1in,height=1.25in,clip]{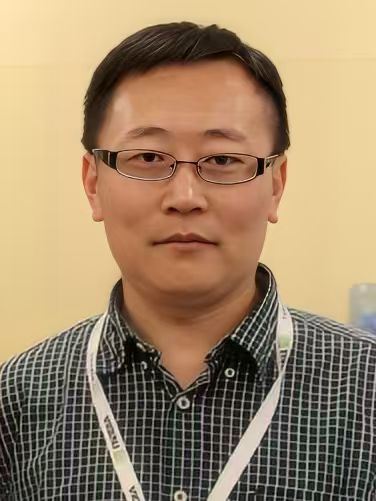}}]
	{Deyu Meng}
	(Member, IEEE) received the BSc, MSc, and PhD degrees in 2001, 2004, and 2008, respectively, from Xi’an Jiaotong University, Xi’an, China. He is currently a professor with School of Mathematics and Statistics, Xi’an Jiaotong University,
	and adjunct professor with Faculty of Information Technology, The Macau University of Science and Technology. He currently serves as an associate editor of IEEE Transactions on Pattern Analysis and Machine Intelligence, Science China-Information Sciences, and Frontiers of Computer Science. His current
	research interests include model-based deep learning, variational networks, and meta-learning.
\end{IEEEbiography}

\begin{IEEEbiography}
	[{\includegraphics[width=1in,height=1.25in,clip]{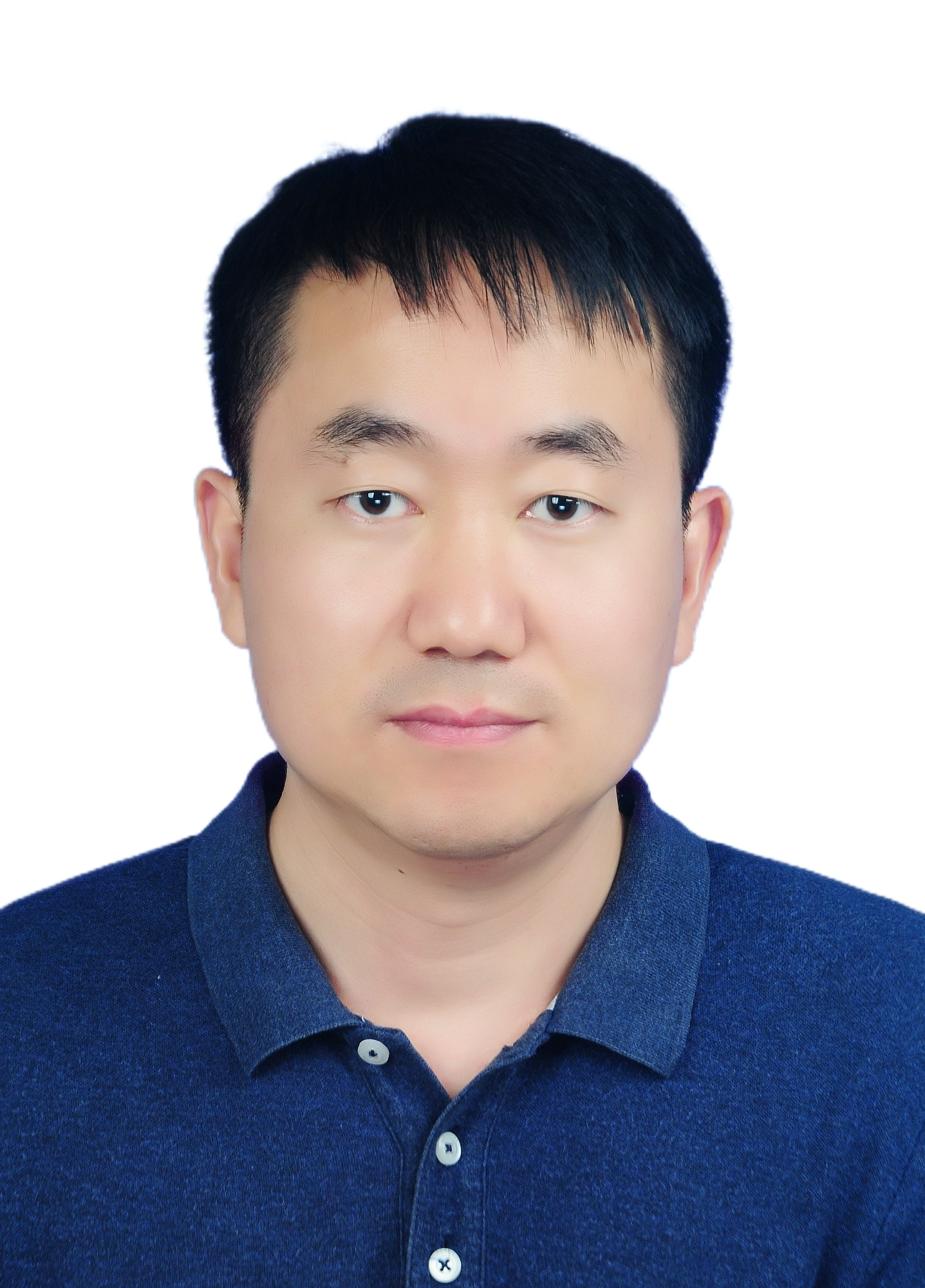}}]
	{Chenqiang Gao} 
	(Member, IEEE) received the B.S. degree in computer science from the China University of Geosciences, Wuhan, China, in 2004 and the Ph.D. degree in control science and engineering from the Huazhong University of Science and Technology, Wuhan, China, in 2009. In August 2009, he joined the School of Communications and Information Engineering, Chongqing University of Posts and Telecommunications (CQUPT), Chongqing, China. In September 2012, he joined the Informedia Group with the School of Computer Science, Carnegie Mellon University, Pittsburgh, PA, USA, working on multimedia event detection (MED) and surveillance event detection (SED) until March 2014, when he returned to CQUPT. In September 2023, he joined the School of Intelligent Systems Engineering, Sun Yat-sen University, Shenzhen, Guangdong, China. His research interests include image processing, infrared target detection, action recognition, and event detection. 
\end{IEEEbiography}

\vfill

\end{document}